%% file: iclr2024_conference.tex
\documentclass{article} % For LaTeX2e
\usepackage{iclr2024_conference,times}

\input{math_commands.tex}

\usepackage{hyperref}
\usepackage{url}

\usepackage{graphicx}
\usepackage{cleveref}
\usepackage{floatrow}
\usepackage{wrapfig}
\usepackage{booktabs, makecell, multirow, tabularx,
            threeparttable, tabulary}

\usepackage{subcaption}

\usepackage[tableposition=top]{caption}

\usepackage{pifont}% http://ctan.org/pkg/pifont

\def\ppiformer{\mbox{\textsc{PPIformer}} }
\def\ppiformern{\mbox{\textsc{PPIformer}}}
\def\ppiref{\mbox{PPIRef} }

\def\ddg{\mbox{$\Delta \Delta G$} }
\def\ddgn{\mbox{$\Delta \Delta G$}}

\def\angstrom{\mathrm{\mathring{A}} }

\usepackage{algorithm}
\usepackage{algpseudocode}

\title{Learning to design protein--protein interactions with enhanced generalization}

\author{
    \centerline{
    Anton Bushuiev$^{1\ast}$\qquad
    Roman Bushuiev$^{1,4}\thanks{These authors contributed equally.}$~~\qquad
    Petr Kouba$^{1,2}$\qquad
    Anatolii Filkin$^{1}$
    } \\ 
    \centerline{\textbf{
    Marketa Gabrielova$^{1}$\qquad
    Michal Gabriel$^{1}$\qquad
    Jiri Sedlar$^{1}$\qquad
    Tomas Pluskal$^{4}$
    }}\\
    \centerline{\textbf{
    Jiri Damborsky$^{2,3}$\qquad
    Stanislav Mazurenko$^{2,3}$\qquad 
    Josef Sivic$^{1}$
    }}\\
    \centerline{$^{1}$Czech Institute of Informatics, Robotics and Cybernetics, Czech Technical University}\\
    \centerline{$^{2}$Loschmidt Laboratories, Department of Experimental Biology and RECETOX, Masaryk University}\\
    \centerline{$^3$International Clinical Research Center, St. Anne's University Hospital Brno}\\
    \centerline{$^4$Institute of Organic Chemistry and Biochemistry of the Czech Academy of Sciences}
}

\iclrfinalcopy % Uncomment for camera[-ready version, but NOT for submission.
\begin{document}

\maketitle

\begin{abstract}
Discovering mutations enhancing protein--protein interactions (PPIs) is critical for advancing biomedical research and developing improved therapeutics. While machine learning approaches have substantially advanced the field, they often struggle to generalize beyond training data in practical scenarios. The contributions of this work are three-fold. First, we construct PPIRef, the largest and non-redundant dataset of 3D protein--protein interactions, enabling effective large-scale learning. Second, we leverage the PPIRef dataset to pre-train PPIformer, a new SE(3)-equivariant model generalizing across diverse protein-binder variants. We fine-tune PPIformer to predict effects of mutations on protein--protein interactions via a thermodynamically motivated adjustment of the pre-training loss function. Finally, we demonstrate the enhanced generalization of our new PPIformer approach by outperforming other state-of-the-art methods on new, non-leaking splits of standard labeled PPI mutational data and independent case studies optimizing a human antibody against SARS-CoV-2 and increasing the thrombolytic activity of staphylokinase.
\end{abstract}

%%%%%%%%%%%%%%%%%%%%%%%%%%%%%%%%%%%%%%%%%%%%%%%%%%%%%
\section{Introduction}
%%%%%%%%%%%%%%%%%%%%%%%%%%%%%%%%%%%%%%%%%%%%%%%%%%%%%

The goal of this work is to develop a reliable method for designing protein--protein interactions (PPIs). We focus on predicting binding affinity changes of protein complexes upon mutations. This problem, also referred to as \ddg prediction, is the central challenge of protein binder design \citep{marchand2022computational}. The discovery of mutations increasing binding affinity unlocks application areas of tremendous importance, most notably in healthcare and biotechnology. Interactions between proteins play a crucial role in mechanisms of various diseases including cancer and neurodegenerative disorders  \citep{lu2020recent, ivanov2013targeting}. They also offer potential pathways for the action of protein-based therapeutics in addressing other medical conditions, such as stroke, which stands as a leading cause of disability and mortality worldwide \citep{feigin2022world, nikitin2022computer}. Furthermore, the design of PPIs is also relevant to biotechnological applications, including the development of biosensors \citep{scheller2018generalized, langan2019novo}.

While machine learning methods for designing protein--protein interactions have been developed for more than a decade, their generalization beyond training data is hindered by multiple challenges. First, training datasets for protein--protein interactions suffer from severe redundancy and biases. These inherent imperfections are difficult to identify and rectify using available algorithmic tools due to the low scalability of the latter. Second, the approaches employed for the train-test splitting of both large-scale unlabeled PPIs and small mutational libraries introduce data leakage, as interactions with near-duplicate 3D structures appear both in the training and test sets. As a result, performance estimates of machine learning models do not accurately reflect their real-world generalization capabilities. Third, commonly employed evaluation metrics do not fully capture practically important performance criteria. Finally, the design and training of the existing models for predicting binding affinity changes upon mutations are often prone to overfitting, as they do not fully capture the right granularity of the protein complex representation and the appropriate inductive biases.

In this work, we make a step towards more generalizable machine learning for the design of protein--protein interactions. First, we construct PPIRef\footnote{\url{https://github.com/anton-bushuiev/PPIRef}}, a comprehensive dataset of protein--protein interaction structures from the Protein Data Bank. Using iDist, a new scalable algorithm, we enhance the dataset quality by clustering PPIs and removing structural redundancy. Second, we introduce \ppiformern\footnote{\url{https://github.com/anton-bushuiev/PPIformer}}\textsuperscript{,}\footnote{\url{https://huggingface.co/spaces/anton-bushuiev/PPIformer}}, an $SE(3)$-equivariant transformer for modeling coarse-grained PPI structures. The transformer is pre-trained on PPIRef to generalize across diverse protein binders. Third, we fine-tune \ppiformer to predict binding affinity changes upon PPI mutations ($\ddgn$), leveraging a thermodynamically motivated loss function. We demonstrate that \ppiformer achieves state-of-the-art performance on multiple test sets, including non-leaking splits of standard mutational data, as well as case studies involving antibody design against SARS-CoV-2 and thrombolytic engineering.

%%%%%%%%%%%%%%%%%%%%%%%%%%%%%%%%%%%%%%%%%%%%%%%%%%%%%
\section{Related work}
%%%%%%%%%%%%%%%%%%%%%%%%%%%%%%%%%%%%%%%%%%%%%%%%%%%%%

\paragraph{Predicting the effects of mutations on protein--protein interactions.}
The task of predicting the effects of mutations on protein--protein interactions measured by \ddg has been studied for more than a decade \citep{geng2019finding}. Traditionally, \ddg predictors relied on physics-based simulations and statistical potentials \citep{barlow2018flex, xiong2017bindprofx, dehouck2013beatmusic, schymkowitz2005foldx}. In contrast, more recent machine learning approaches \citep{rodrigues2021mmcsm, pahari2020saambe, wang2020topology, geng2019isee} primarily rely on handcrafted descriptors of protein--protein interfaces. The latest generation of methods employs end-to-end deep learning on protein complexes, showing to surpass computationally intensive force field simulations in terms of predictive performance \citep{luo2023rotamer, shan2022deep, liu2021deep}. In this study, we revisit this claim and show that the reported performance of the state-of-the-art methods may be overestimated due to leaks in the evaluation data and overfitting.

\paragraph{Self-supervised learning for protein design.}
Self-supervised learning provides means to mitigate the high cost of collecting annotated protein mutation data for supervised learning. These methods, including protein language models \citep{meier2021language}, inverse folding \citep{hsu2022learning}, and prediction of missing amino acid atoms \citep{shroff2019structure}, leverage unannotated data to predict the effects of protein mutations. Recent advancements have also demonstrated that self-supervised pre-training on synthetic tasks enhances performance in various tasks related to protein structures \citep{zhang2023enhancing}. Some of the latest binding \ddg predictors utilize pre-training from protein structures by learning to reconstruct native side-chain rotamer angles \citep{luo2023rotamer, liu2021deep}. Our work introduces a model pre-trained on 3D protein interaction structures aimed for the design of protein binders.

\paragraph{Datasets of protein--protein interactions.}

\input{assets/datasets_table}

Historically, datasets of protein--protein interactions have been falling under two categories: sets of hundreds of small curated task-specific examples \citep{jankauskaite2019skempi, vreven2015updates}, and larger collections of thousands of unannotated interactions, potentially containing biases and structural redundancy \citep{morehead2021dips, evans2021protein, townshend2019end, gainza2020deciphering}. In this work, we construct a new dataset that is one order of magnitude larger than existing alternatives from the second category. We further refine our dataset by removing structurally near-duplicate entries using our new scalable 3D structure-matching algorithm. This results in the largest available and non-redundant dataset of protein--protein interaction structures. Our algorithm for comparing protein--protein interactions contrasts with existing methods that rely on computationally expensive alignment procedures \citep{shin2023quantitative, mirabello2018topology, cheng2015pcalign, gao2010ialign}. Instead, we design our algorithm to enable large-scale retrieval of similar protein--protein interfaces by approximating their structural alignment. Our approach complements prior work on the efficient retrieval of similar protein sequences \citep{steinegger2017mmseqs2} and, more recently, monomeric protein structures \citep{van2023fast}.

%%%%%%%%%%%%%%%%%%%%%%%%%%%%%%%%%%%%%%%%%%%%%%%%%%%%%
\section{PPIRef: New large dataset of protein--protein interactions}
%%%%%%%%%%%%%%%%%%%%%%%%%%%%%%%%%%%%%%%%%%%%%%%%%%%%%

The Protein Data Bank (PDB) is a massive resource of over 200,000 experimentally obtained protein 3D structures \citep{berman2000protein}. The space of protein--protein interactions in PDB is hypothesized to cover nearly all physically plausible interfaces in terms of geometric similarity \citep{gao2010structural}. Nevertheless, it comes at the expense of a heavy structural redundancy given by the highly modular anatomy of many proteins and their complexes \citep{draizen2022prop3d, burra2009global}. To the best of our knowledge, there have been no attempts to quantitatively assess the redundancy of the large protein--protein interaction space represented in PDB and construct a balanced subset suitable for large-scale learning. We start this section by introducing the approximate iDist algorithm enabling fast detection of near duplicate protein--protein interfaces (\Cref{sec:idist}). Using iDist, we assess the effective size and splits of existing PPI datasets (\Cref{sec:limits_ppi}) and propose a new, largest and non-redundant PPI dataset, called PPIRef (\Cref{sec:ppiref}).

\subsection{iDist: new efficient approach for protein--protein interface deduplication}\label{sec:idist}

\input{assets/duplicates_example}

Existing algorithms to determine the similarity between protein--protein interfaces rely on structural alignment procedures \citep{gao2010ialign, mirabello2018topology, shin2023ppisurfer}. However, since finding an optimal alignment between PPIs is computationally heavy, alignment-based approaches do not scale to large datasets. Therefore, we develop a reliable approximation of the well-established algorithm iAlign, which adapts the well-known TM-score to the domain of PPIs \citep{gao2010ialign}. Our iDist algorithm calculates $SE(3)$-invariant vector representations of protein--protein interfaces via message passing between residues, which in turn enables efficient detection of near-duplicates using the Euclidean distance with a threshold estimated to approximate iAlign (\Cref{fig:idist_vs_align}). The complete algorithm is described in \Cref{sec:ppiref-appendix}.

To evaluate the performance of our iDist, we benchmark it against iAlign. We start by sampling 100 PDB codes from the DIPS dataset \citep{townshend2019end} of protein--protein interactions and extract the corresponding 1,646 PPIs. Subsequently, we calculate all 2,709,316 pairwise similarities between these PPIs using both the exact iAlign structural alignment algorithm and our efficient iDist approximation. Employing 128 CPUs in parallel, iAlign computations took 2 hours, while iDist required 15 seconds, being around 480 times faster. Next, we estimate the quality of the approximation on the task of retrieving near-duplicate PPI interfaces. Using iAlign-defined ground-truth duplicates, iDist demonstrates a precision of 99\% and a recall of 97\%. These evaluation results confirm that iDist enables efficient structural deduplication of extensive PPI data within a reasonable timeframe, which is not achievable with other existing algorithms. The scalability of the method enabled us to analyze existing PPI datasets and their respective data splits used by the recent machine-learning approaches. The analysis is described below. Please refer to \Cref{sec:ppiref-appendix} for the additional details of the evaluation of iDist.

\subsection{Limitations of existing protein--protein interaction datasets}\label{sec:limits_ppi}

We apply iDist to assess the composition of DIPS -- the state-of-the-art dataset of protein--protein interactions comprising approximately 40,000 entries \citep{morehead2021dips, townshend2019end}. We construct a near-duplicate graph of DIPS by connecting two protein--protein interfaces if they are detected as near duplicates by iDist (see \Cref{fig:dips_duplicates} for an example). This results in a graph with 8.5K connected components, where the largest connected component comprises 36\% of the interfaces. Notably, relaxing iDist duplicate detection threshold 1.5 times results in 84\% of the interfaces forming a single component, indicating the high connectivity of the PPI space in DIPS. After iteratively deduplicating entries with at least one near duplicate, the dataset size drops to 22\% of its initial size. These observations are in agreement with the hypothesis of \cite{gao2010structural} suggesting high connectivity and redundancy of the PPI space in PDB.

Finally, we analyze DIPS data splits, estimating leakage by the ratio of test interactions with near duplicates in training or validation folds. We find that the split based on protein sequence similarity (not 3D structure as in our case) used for the validation of protein--protein docking models \citep{ketata2023diffdock-pp, ganea2021independent} has near duplicates in the training data for 53\% of test examples, while the random split from \cite{morehead2021dips} has 88\% of test examples with a near duplicate in the training data. \Cref{fig:dips_duplicates} illustrates an example of such a leak in the original split \citep{ganea2021independent}: the interface from the test fold (left) and a near-duplicate from the training fold (right). In \cite{bushuiev2024revealing}, we demonstrate that high data leakage is present in most splits of PPI structures across multiple tasks due to the insufficiency of similarity measures commonly employed to test for data leakage and create the splits.

\subsection{PPIRef: New large dataset of protein--protein interactions}\label{sec:ppiref}

We address the redundancy of existing PPI datasets by building a novel dataset of structurally distinct 3D protein--protein interfaces, which we call PPIRef. We start by exhaustively mining all 202,380 entries from the Protein Data Bank as of June 20, 2023. Subsequently, we extract all putative interactions by finding all pairs of protein chains that have a contact between heavy atoms in the range of at most $10\angstrom$. This procedure results in 837,241 hypothetical PPIs, further referred to as the raw PPIRef800K. Further, we apply the well-established criteria (\Cref{sec:ppiref-appendix}, \cite{townshend2019end}) to select only biophysically proper interactions. This filtering results in 322,454 protein--protein interactions comprising the vanilla version of PPIRef, which we name PPIRef300K. Finally, we iteratively filter out near-duplicate entries by applying the iDist algorithm. This deduplication results in the final non-redundant dataset comprising 45,553 PPIs, which we call PPIRef50K (or simply PPIRef). \Cref{tab:ppi_data_stats} illustrates that our dataset exceeds the total and effective sizes of the representative alternatives: DIPS \citep{townshend2019end}, DIPS-Plus \citep{morehead2021dips}, and the dataset used to train the MaSIF-search model \citep{gainza2020deciphering}, also used by the recent MaSIF-seed pipeline for \textit{de novo} PPI design \citep{gainza2023novo}.

%%%%%%%%%%%%%%%%%%%%%%%%%%%%%%%%%%%%%%%%%%%%%%%%%%%%%
\section{Learning to design protein--protein interactions}
%%%%%%%%%%%%%%%%%%%%%%%%%%%%%%%%%%%%%%%%%%%%%%%%%%%%%

\begin{figure}[t!]
  \centering
  \includegraphics[width=\textwidth]{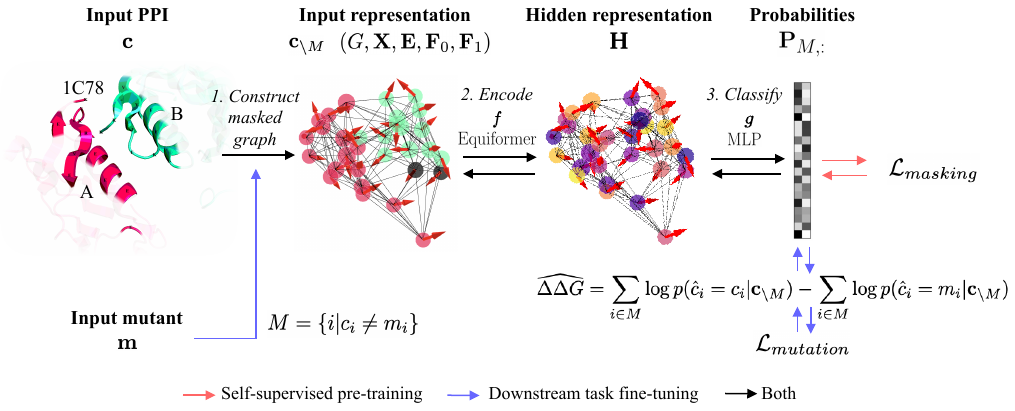}
  \vspace*{-5mm}
  \caption[Overview of \ppiformer.]{Overview of \ppiformern. A single pre-training step starts with randomly sampling a protein--protein interaction $\mathbf{c}$ (in this example, the staphylokinase dimer A--B from the PDB entry 1C78) from PPIRef. Next, randomly selected residues $M$ are masked to obtain the masked interaction $\mathbf{c}_{\setminus M}$. After that, the interaction is converted into a graph representation $(G, \mathbf{X}, \mathbf{E}, \mathbf{F}_0, \mathbf{F}_1)$ with masked nodes $M$ (black circles). The model subsequently learns to classify the types of masked amino acids by acquiring $SE(3)$-invariant hidden representation $\mathbf{H}$ of the whole interface via the encoder $f$ and classifier $g$ (red arrows). On the downstream task of \ddg prediction, mutated amino acids are masked, and the probabilities of possible substitutions $\mathbf{P}_{M,:}$ are jointly inferred with the pre-trained model. Finally, the estimate $\widehat{\Delta \Delta G}$ is obtained using the predicted probabilities $p$ of the wild-type $c_i$ and the mutant $m_i$ amino acids via log odds (blue arrows).}
  \label{fig:ppiformer}
\end{figure}

Predicting how amino-acid substitutions affect binding energies of protein complexes (\ddgn) is crucial for designing protein interactions but is hindered by limited data, covering just a few hundred interactions \citep{jankauskaite2019skempi}. This section presents our approach, including developing coarse-grained representations of protein complexes to avoid overfitting, introducing the $SE(3)$-equivariant \ppiformer model to learn from the coarse-grained representations, and detailing our structural masked modeling for pre-training on the large PPIRef dataset. Finally, we present how we fine-tune \ppiformer to predict \ddg values using a thermodynamically motivated loss function.

\subsection{Representation of protein--protein complexes}\label{subsec:representation}

In a living cell, proteins continually undergo thermal fluctuations and change their precise shapes. This fact is particularly manifested on protein surfaces, where the precise atomic structure is flexible due to interactions with other molecules. As a result, protein--protein interfaces may be highly flexible \citep{kastritis2013binding}. Nevertheless, the available crystal structures from the Protein Data Bank only represent their rigid, energetically favorable states \citep{jin2023unsupervised}. Therefore, we aim to develop representations of protein complexes robust to atom fluctuations, as well as well-suited for modeling mutated interface variants. In this section, we define a coarse residue-level representation to allow for sufficient flexibility of the interfaces, which nevertheless captures the major aspects of the interaction patterns.

More specifically, consider a protein--protein complex (or interface) $\mathbf{c} \in \mathbb{\mathcal{A}}^N$ of $N$ residues in the alphabet of amino acids $\mathcal{A} = \{1, \dots, 20\}$. We consider the order of residues in $\mathbf{c}$ arbitrary, ensuring permutation invariance, a property critical for modeling the interfaces \citep{mirabello2018topology}. Next, we represent the complex $\mathbf{c}$ as a $k$-NN graph $G$, where the nodes represent the individual residues and edges are based on the proximity of corresponding alpha-carbon ($C_{\alpha}$) atoms. The graph is augmented by node-level and pair-wise features $\mathbf{X}, \mathbf{E}, \mathbf{F}_0, \mathbf{F}_1$. In detail, matrix $\mathbf{X} \in \mathbb{R}^{N \times 3}$ contains the coordinates of alpha-carbons of all residues. Next, all residue nodes are put into semantic relation by pair-wise binary edge features $\mathbf{E} \in \{0, 1\}^{N \times N}$ equal to 0 if residues come from the same protein partner and 1 otherwise. Finally, each node is associated with two kinds of features: type-0 $\mathbf{F}_0 \in \mathbb{R}^{N \times 20 \times 1}$, also referred to as scalars, and type-1 vectors $\mathbf{F}_1 \in \mathbb{R}^{N \times 1 \times 3}$. Features $\mathbf{F}_0$ capture the one-hot representations of wild-type amino acids $\mathbf{c}$, while vectors $\mathbf{F}_1$ are defined as virtual beta-carbon orientations calculated from the backbone geometry using ideal angle and bond length definitions \citep{dauparas2022robust}. 

Please note that $\mathbf{F}_0$ are invariant with respect to rototranslations and $\mathbf{F_1}$ are equivariant. Collectively, coordinates $\mathbf{X}$ and virtual-beta carbon directions $\mathbf{F_1}$ capture the complete geometry of the protein backbones involved in the complex. Our representation is agnostic to precise angles of side-chain rotamers, implicitly modeling their flexibility. The schematic illustration is provided in \Cref{fig:ppiformer}.

\subsection{\textsc{PPIformer} model}\label{subsec:ppiformer}

In order to effectively learn from protein--protein complexes or interfaces $(G, \mathbf{X}, \mathbf{E}, \mathbf{F}_0, \mathbf{F}_1)$, i.e.~respecting permutation invariance of amino acids and arbitrary coordinate systems of Protein Data Bank entries, we define \ppiformern, an $SE(3)$-equivariant architecture. The model consists of an encoder $f$ and classifier $g$ such that $g(f(G, \mathbf{X}, \mathbf{E}, \mathbf{F}_0, \mathbf{F}_1))$ yields a probability matrix $\mathbf{P} \in [0,1]^{N \times |\mathcal{A}|}$, where $ P_{i,j}$ defines the probability of amino-acid type $j$ at residue $i$. Intuitively, matrix $\mathbf{P}$ represents amino acid probabilities in a protein complex based on its structure.

The core of our architecture is comprised of Equiformer $SE(3)$-equivariant graph attention blocks $f^{(l)}$ \citep{liao2023equiformerv2, liao2022equiformer}. Each of the $L$ blocks (or layers) updates equivariant features of different types associated with all amino acids via message passing with an equivariant attention mechanism with $K$ heads. In detail, the input to the $l$\textsuperscript{th} block is an original graph $G$ with coordinates $\mathbf{X}$ and edge features $\mathbf{E}$ along with a set of node feature matrices $\mathbf{H}_{k}^{(l)} \in \mathbb{R}^{N \times d_{k} \times (2k+1)}$ of different equivariance types $k \geq 0$ from the previous layer. Here, $d_k \geq 0$ is a hyper-parameter defining the number of type-$k$ hidden features, which is shared across all blocks, and $2k+1$ is their corresponding theoretical dimension. The input node features for the first layer are set to $\mathbf{F}_0, \mathbf{F}_1$. The output of each block is given by the updated node features of different equivariance types. Internally, all blocks lift hidden features up to the equivariant representations of the degree $\deg \geq 0$ which is an additional hyper-parameter. Taking the type-0 outputs of the final layer leads to invariant amino acid embeddings $\mathbf{H}$ as:
\begin{align}
    \mathbf{H}^{(1)}_0, \mathbf{H}^{(1)}_1, \dots, \mathbf{H}^{(1)}_{deg} &= f^{(0)}(G, \mathbf{X}, \mathbf{E}, \mathbf{F}_0, \mathbf{F}_1), \\
    \mathbf{H}^{(l+1)}_0, \mathbf{H}^{(l+1)}_1, \dots, \mathbf{H}^{(l+1)}_{deg} &= f^{(l)}(G, \mathbf{X}, \mathbf{E}, \mathbf{H}^{(l)}_0, \mathbf{H}^{(l)}_1, \dots, \mathbf{H}^{(l)}_{deg}), \\
    \mathbf{H} &:= \mathbf{H}^{(L-1)}_0.
\end{align}
Collectively, we term the composition of transformer blocks $f^{(l)}$ (for $l \in \{ 0, \dots, L-1 \}$) as the encoder
\begin{align}
    f &: G, \mathbf{X}, \mathbf{E}, \mathbf{F}_0, \mathbf{F}_1  \mapsto \mathbf{H}
\end{align}
with the property of $SE(3)$-invariance for any rotation $\mathbf{R} \in SO(3)$ and translation $\mathbf{t} \in \mathbb{R}^{3}$:
\begin{align}
    f(G, \mathbf{X}, \mathbf{E}, \mathbf{F}_0, \mathbf{F}_1) &= f(G, \mathbf{X}\mathbf{R} + \mathbf{1}\mathbf{t}^T, \mathbf{E}, \mathbf{F}_0, \mathbf{F}_1 \mathbf{R}).
\end{align}
To further estimate the probabilities of masked amino acids discussed below, we apply a 1-layer classifier with the softmax activation $g: \mathbb{R}^{N \times d_0 \times 1} \to [0,1]^{N \times \mathcal{|A|}}$ on top of node embeddings $\mathbf{H}$ to obtain the probability matrix $\mathbf{P}$. The illustration of the model is provided in \Cref{fig:ppiformer} with the described pipeline depicted with black arrows.

\subsection{3D equivariant self-supervised pre-training from unlabeled protein--protein interactions}\label{subsec:ssl}

In this section, we describe how we leverage a large amount of unlabeled protein--protein interfaces from \ppiref to train \ppiformer to capture the space of native PPIs via masked modeling.

\paragraph{Structural masking of protein--protein interfaces.} The paradigm of masked modeling has proven to be an effective way of pre-training from protein sequences \citep{lin2023evolutionary}. Nevertheless, while the masking of amino acids in a protein sequence is straightforward by introducing a special token, the masking of structural fragments is not obvious. Here, we leverage our flexible coarse-grained protein--protein complex representation to define masking by a simple change in the feature representation.

Having a protein complex or interface $\mathbf{c} \in \mathcal{A}^{N}$ containing $N$ amino acids from vocabulary $\mathcal{A}$ and a mask $M \subset \{1, \dots, N\}$, we define the masked structure $\mathbf{c}_{\setminus M} \in (\mathbb{\mathcal{A}} \cup \{0\})^N$ by setting amino acid classes at all masked positions $M$ to zeros. Consequently, when constructing one-hot scalar features $\mathbf{F}_0$ from $\mathbf{c}_{\setminus M}$, we set the corresponding rows to zeroes. Note that vector features $\mathbf{F}_1$ do not require masking since they do not contain any information about the types of amino acids, benefiting from using virtual beta-carbons instead of real ones. Additionally, the glycine amino acid, which lacks the beta-carbon atom, does not need special handling.

\paragraph{Loss for masked modeling of protein--protein interfaces.} Having a protein--protein interaction $\mathbf{c} \in \mathbb{\mathcal{A}}^N$ and a random mask $M \subset \{1, \dots, N\}$, we use a traditional cross-entropy loss for training the model to predict native amino acids in a masked version $\mathbf{c}_{\setminus M} \in (\mathbb{\mathcal{A}} \cup \{0\})^N$ of the native interaction. To additionally increase the generalization capabilities of \ppiformer to capture potentially unseen or mutated interfaces, we further employ two regularization measures. First, we apply label smoothing \citep{szegedy2016rethinking} to force the model not to be overly confident in native amino acids, so that it can be more flexible towards unseen variants. Second, we weight the loss inversely to the prior distribution of amino acids in protein--protein interfaces to remove the bias towards overrepresented amino acids such as leucine. The overall pre-training loss is defined as
\begin{align}\label{eq:loss-ce}
    \mathcal{L}_{masking} &= - \sum_{i \in M}{w_{c_i} \Biggl[ (1 - \epsilon) \cdot \log{p(\hat{c}_i = c_i \vert \mathbf{c}_{\setminus M})} + \epsilon\sum_{j \in \mathcal{A} \setminus c_i}{\log{p(\hat{c}_i = c_j \vert \mathbf{c}_{\setminus M})} \cdot \frac{1}{|\mathcal{A}|} \Biggr]}}.
\end{align}
Here, $p(\hat{c}_i = c_i \vert \mathbf{c}_{\setminus M}) := P_{i,c_i}$ is the probability of predicting the native amino acid type $c_i$ of a masked residue $i \in M$. Next, the sum over all other, non-native, amino acid types $j \in \mathcal{A} \setminus c_i$ is the label smoothing regularization term where $0 < \epsilon < 1$ is the smoothing hyper-parameter. Further, $w_{c_i}$ is the weighting factor corresponding to the native amino acid $c_i$. Finally, the sum over $i \in M$ corresponds to the loss being evaluated over all masked residues.

\subsection{Transfer learning for predicting the effects of mutations on protein--protein interactions}\label{subsec:fine-tuning}

Predicting the effects of mutations on binding affinity (\ddgn) is a central task in designing protein--protein interactions. Nevertheless, collecting  $\Delta \Delta G$ annotations is expensive and time-consuming. As a result, the labeled mutational data for binding affinity changes are scarce, not exceeding several thousand annotations \citep{jankauskaite2019skempi}. Therefore, in our work, we aim to leverage the pre-trained \ppiformer model for scoring mutations with minimal supervision.

\paragraph{Thermodynamic motivation.} From the thermodynamic perspective, binding energy change \ddg (or alternatively denoted as $\Delta \Delta G_{wt \rightarrow mut}$) can be decomposed as follows:
\begin{align}\label{eq:ddg}
    \Delta \Delta G &= \Delta G_{mut} - \Delta G_{wt} = RT(\log{(K_{wt})} - \log{(K_{mut})}),
\end{align}
where $\Delta G_{mut}$ and $\Delta G_{wt}$ are binding energies of mutated and wild-type complexes, respectively \citep{kastritis2013binding}, $R, T > 0$ are the gas constant and temperature of the environment, respectively, and $K_{wt}$ and $K_{mut}$ denote the equilibrium constants of protein--protein interactions, i.e.~the ratios of the concentration of the complexes formed when proteins interact to the concentrations of the non-interacting proteins. The form of $\Delta \Delta G$ (\Cref{eq:ddg}) introduces symmetries into the problem of estimating the quantity. For example, predicting the effect of a reversed mutation $\Delta \Delta G_{mut \rightarrow wt}$ should satisfy the antisymmetry property $\Delta \Delta G_{wt \rightarrow mut} = - \Delta \Delta G_{mut \rightarrow wt}$. Available machine learning predictors either ignore the symmetry \citep{liu2021deep} or require two forward passes to estimate the quantity twice, for both directions, and combine the predictions to enforce the antisymmetry as ($\Delta \Delta G_{wt \rightarrow mut} - (- \Delta \Delta G_{mut \rightarrow wt}) )/ 2$ \citep{luo2023rotamer}.

\paragraph{Predicting the effects of mutations on binding energy via the log odds ratio.} Here, in line with physics-informed machine learning \citep{karniadakis2021physics}, we leverage the thermodynamic interpretation of \ddg to adapt the pre-training cross-entropy for the downstream \ddg fine-tuning. Having a complex $\mathbf{c} \in \mathbb{\mathcal{A}}^N$ and its mutant $\mathbf{m} \in \mathbb{\mathcal{A}}^N$, with the substitutions of residues $M \subset \{1, \dots, N\}$  such that $c_i \neq m_i$ for all $i \in M$, we estimate its binding energy change as:
\begin{align}\label{eq:ddg-pred}
    \widehat{\Delta \Delta G} &= \sum_{i \in M}{\log{p(\hat{c}_i = c_i \vert \mathbf{c}_{\setminus M})}} - \sum_{i \in M}{\log{p(\hat{c}_i = m_i \vert \mathbf{c}_{\setminus M})}},
\end{align}
where the $p$ terms are \ppiformer output probabilities. The $\widehat{\Delta \Delta G}$ prediction is the log odds ratio, used by \cite{meier2021language} for zero-shot predictions on protein sequences. Intuitively, the predicted binding energy change upon mutation is negative (increased affinity) if the predicted likelihood of the mutated structure is higher than the likelihood of the native structure. When simultaneously decomposing \Cref{eq:ddg} and \Cref{eq:ddg-pred}, we observe that $\log{(K_{wt})}$ is estimated as $\sum_{i \in M}{\log{p(\hat{c}_i = c_i \vert \mathbf{c}_{\setminus M})}}$, and $\log{(K_{mut})}$ is estimated as $\sum_{i \in M}{\log{p(\hat{c}_i = m_i \vert \mathbf{c}_{\setminus M})}}$. Considering that the estimate of the wild-type likelihood is identical to the pre-training loss (\Cref{eq:loss-ce}) up to the regularizations, we reason that during pre-training \ppiformer learns the correlates of $\Delta G$ values, whereas during fine-tuning it refines them to $\ddg$ predictions. For fine-tuning through \Cref{eq:ddg-pred}, we use the MSE loss denoted as $\mathcal{L}_{mutation}$.

%%%%%%%%%%%%%%%%%%%%%%%%%%%%%%%%%%%%%%%%%%%%%%%%%%%%%
\section{Experiments}
%%%%%%%%%%%%%%%%%%%%%%%%%%%%%%%%%%%%%%%%%%%%%%%%%%%%%

In this section, we describe our protocol for benchmarking the generalization on the task of \ddg prediction and present our results. We begin by introducing the evaluation datasets, metrics, and baseline methods (\Cref{sec:eval_protocol}). Next, we show that our approach outperforms state-of-the-art machine learning methods in designing protein--protein interactions distinct from the training data (\Cref{sec:comparison-with-sota}). Additionally, we show the benefits of our new PPIRef dataset and key \ppiformer components through ablation studies in \Cref{sec:appendix-ablations}.

\subsection{Evaluation protocol}\label{sec:eval_protocol}

\paragraph{Datasets.}

To fine-tune \ppiformer for \ddg prediction we use the largest available labeled dataset, SKEMPI v2.0, containing 7085 mutations \citep{jankauskaite2019skempi}. Prior works \citep{luo2023rotamer, liu2021deep} primarily validate the performance of models on PDB-disjoint splits of the dataset. However, we find such approach not appropriate to measure the generalization capacity of predictors due to a high ratio of leakages. Therefore, we construct a new, non-leaking cross-validation split and set aside 5 PPI outliers to obtain 5 additional distinct test folds. Further, to simulate practical protein--protein interaction design scenarios, we perform additional evaluation on two independent case studies. These case studies assess the capability of models to retrieve mutations optimizing the human P36-5D2 antibody against SARS-CoV-2 and increasing the thrombolytic activity of staphylokinase. Please refer to \Cref{sec:eval-data} for details.

\paragraph{Metrics.}

To evaluate the capabilities of models in prioritizing favorable mutations, we use the Spearman correlation coefficient between the predicted and ground-truth \ddg values. To evaluate the performance in detecting stabilizing mutations, we calculate precision and recall with respect to negative \ddg values. We also report additional metrics to enable comparison with prior work \citep{luo2023rotamer}. However, in \Cref{sec:eval-metrics}, we emphasize that these metrics can be misleading when selecting a model for a practical application. For the independent retrieval case studies, we report the predicted ranks for all favorable mutations and evaluate the retrieval performance using the precision at $k$ metrics (P@$k$). We provide the details in \Cref{sec:eval-metrics}.

\paragraph{Baselines.}

We evaluate the performance of \ppiformer against the state-of-the-art represenatives in 5 main categories of available methods. Specifically, the flex ddG \citep{barlow2018flex} Rosetta-based \citep{leman2020rosetta} protocol is the most advanced traditional force-field simulator. GEMME is a state-of-the-art non-learning method based on evolutionary trees of sequences \citep{laine2019gemme}. The baseline machine learning methods are as follows: supervised RDE-Network \citep{luo2023rotamer}, pre-trained on unlabeled protein structures and fine-tuned for \ddg prediction on SKEMPI v2.0; ESM-IF \citep{hsu2022learning}, an unsupervised predictor of \ddg trained for inverse folding on experimental and \mbox{AlphaFold2} \citep{jumper2021highly} structures; and MSA Transformer \citep{rao2021msa}, an evolutionary baseline. Please see \Cref{sec:baselines} for details.

\subsection{Comparison with the state of the art}\label{sec:comparison-with-sota}

\paragraph{Prediction of \ddg on held-out test cases.}

As shown in \Cref{fig:skempi_test}, on 5 challenging test PPIs from SKEMPI v2.0, \ppiformer confidently outperforms all machine-learning baselines in 6 out of 7 evaluation metrics, being the second-best in terms of recall. We achieve a 183\% relative improvement in mutation ranking compared to the state-of-the-art supervised RDE-Network, as measured by Spearman correlation. Importantly, this non-leaking evaluation reveals that traditional force field simulators, represented by state-of-the-art flex ddG, still outperform machine learning methods in terms of predictive performance. However, in terms of speed, they may not be applicable in typical real-world mutational screenings, being 5 orders of magnitude slower (see \Cref{sec:baselines}).

\input{assets/table_skempi_test_overall}

\paragraph{Optimization of a human antibody against SARS-CoV-2.}
Within a pool of 494 candidate single-point mutations of a human antibody, our model detects 2 out of 5 annotated mutations (using the top-10\% threshold as defined by \cite{luo2023rotamer}) that are known to be effective against SARS-CoV-2 (\Cref{tab:res-covid}). The best among the other methods detect 3 out of 5 mutations. However, in contrast to the other methods, \ppiformer successfully assigns the best rank to one of the 5 favorable mutations (P@1 = 100\%). Besides that, \ppiformer achieves superior performance when considering the ranks of all 5 mutations collectively, with the maximum rank for a favorable mutation not exceeding 21.46\% and the mean rank of 11.60\% (compared to the second best values of 51.42\% and 18.26\%, respectively). Overall, \ppiformer is superior in prioritizing favorable mutations among random ones but does not prioritize 5 annotated mutants as high as the DDGPred and RDE-Network models. This suggests that there may be other, potentially more favorable, candidates in the unannotated pool of other 489 out of 494 single-point mutations.

\input{assets/table_ranking_covid}

\paragraph{Engineering staphylokinase for enhanced thrombolytic activity.}

Within a pool of 80 annotated mutations, \ppiformer precisely prioritizes 2 out of 6 strongly favorable ones (i.e., increasing the thrombolytic activity at least twice) as the top-2 mutation candidates (\Cref{tab:sak}). In contrast, the second-best method, RDE-Network, identifies one mutation and assigns it a worse ranking (top-4). MSA Transformer provides accurate top-1 prediction but strongly loses performance on higher cutoffs and detecting two-fold activity-increasing mutations.

\input{assets/table_ranking_sak}

\section{Conclusion}

In this work, we have constructed PPIRef -- the largest and non-redundant dataset for self-supervised learning from protein--protein interactions. Using this data, we have pre-trained a new model, \ppiformern, designed to capture diverse protein binding modes. Subsequently, we fine-tuned the model for the target task of discovering mutations enhancing protein--protein interactions. We have shown that our model effectively generalizes to unseen PPIs, outperforming other state-of-the-art machine learning methods. This work opens up the possibility of training large-scale foundation models for protein--protein interactions.

\subsubsection*{Acknowledgments}

This work was supported by the Ministry of Education, Youth and Sports of the Czech Republic through projects e-INFRA CZ [ID:90254], ELIXIR [LM2023055], CETOCOEN Excellence CZ.02.1.01/0.0/0.0/17\_043/0009632, ESFRI RECETOX RI LM2023069.  This work was also supported by the European Union (ERC project FRONTIER no. 101097822) and the CETOCOEN EXCELLENCE Teaming project supported from the European Union’s Horizon 2020 research and innovation programme under grant agreement No 857560. This work was also supported by the Czech Science Foundation (GA CR) through grant 21-11563M and through the European Union’s Horizon 2020 research and innovation programme under Marie Skłodowska-Curie grant agreement No.~891397. Views and opinions expressed are however those of the author(s) only and do not necessarily reflect those of the European Union or the European Research Council. Neither the European Union nor the granting authority can be held responsible for them. Petr Kouba is a holder of the Brno Ph.D. Talent scholarship funded by the Brno City Municipality and the JCMM. We thank David Lacko for preparing the dataset of staphylokinase mutations.

\bibliography{iclr2024_conference}
\bibliographystyle{iclr2024_conference}

\clearpage
\section*{Appendix}

The first section of the appendix provides additional details on the evaluation of the iDist algorithm by benchmarking it against the iAlign structural alignment method, as well as the details of the PPIRef dataset (\Cref{appendix-data}). Next, we provide the implementation (\Cref{appendix-implementation}) and experimental (\Cref{appendix-experimental-details}) details, including the selection of training hyper-parameters, the construction of the mutation datasets, and the description of the evaluation metrics and baselines. Next, we discuss several ablation studies illustrating the importance of the proposed \ppiformer components (\Cref{sec:appendix-ablations}). Finally, we provide the details of the comparison of our method with the state-of-the-art $\Delta \Delta G$ predictors on the SKEMPI v2.0 test set~(\Cref{appendix-additional-results}).

\appendix
\section{Details of the iDist  algorithm and PPIRef dataset}\label{sec:ppiref-appendix}\label{appendix-data}

In this section, we discuss the details relevant to the analysis and construction of large unannoted PPI datasets. First, we provide the complete description of the iDist deduplication algorithm and the calibration of its distance threshold. Second, we specify the construction criteria for our PPIRef.

\paragraph{iDist algorithm.}

\begin{algorithm}[H]
\centering
\caption{
    iDistEMBED
}
\label{alg:embed}
\begin{algorithmic}[1]
    \State{\textbf{Input:} Protein--protein interface $\mathcal{I}$ of $N$ residues from $C$ chains.}
    \State{\textbf{Output:} Vector representation of the interface $\mathbf{z}_\mathcal{I}$.}
    \State{// Get coordinates, features, and partner information}
    \State{$\mathbf{X} \in \mathbb{R}^{N \times 3}, \mathbf{F} \in \mathbb{R}^{N \times d}, \mathbf{p} \in \{1,\dots, C\}^N \gets get\_residues(\mathcal{I})$}
    \State{// Embed residues}
    \For{$i \gets 1$ to $N$}
        \State{$J_{intra} \gets \{j \in \{1, \dots, N\} \mid p_i = p_j \}$}
        \State{$J_{inter} \gets \{j \in \{1, \dots, N\} \mid p_i \neq p_j \}$}
        \State{$\mathbf{m}_{intra} \gets \frac{1}{|J_{intra}|} \sum_{j \in J_{intra}}{\mathbf{f}_j \cdot e^{-\frac{\|\mathbf{x}_i - \mathbf{x}_j\|_{2}^{2}}{\alpha}}}$}
        \State{$\mathbf{m}_{inter} \gets \frac{1}{|J_{inter}|} \sum_{j \in J_{inter}}{\mathbf{f}_j \cdot e^{-\frac{\|\mathbf{x}_i - \mathbf{x}_j\|_{2}^{2}}{\alpha}}}$}
        \State{$\mathbf{h}_i \gets \frac{1}{2}\mathbf{f}_i + \frac{1}{4}\mathbf{m}_{intra} - \frac{1}{4}\mathbf{m}_{inter}$}
    \EndFor
    \State{// Embed interface}
    \For{$c \gets 1$ to $C$}
        \State{$J_{c} \gets \{j \in \{1, \dots, N\} \mid p_j = c\}$}
    \EndFor
    \State{$\mathbf{z}_\mathcal{I} \gets \frac{1}{|C|} \sum_{c=1}^{C}{\frac{1}{|J_c|}\sum_{j \in J_c}{\mathbf{h}_j}}$}
    \State \textbf{return} $\mathbf{z}_\mathcal{I}$
\end{algorithmic}
\end{algorithm}

\begin{algorithm}[H]
\centering
\caption{
    iDist
}
\label{alg:compare}
\begin{algorithmic}[1]
    \State{\textbf{Input:} Two protein--protein interfaces $\mathcal{I}$ and $\mathcal{J}$.}
    \State{\textbf{Output:} Distance $\geq 0$.}
    \State{$\mathbf{z}_\mathcal{I} \gets \text{iDistEMBED}(\mathcal{I})$}
    \State{$\mathbf{z}_\mathcal{J} \gets \text{iDistEMBED}(\mathcal{J})$}
    \State \textbf{return} $\|\mathbf{z}_\mathcal{I} - \mathbf{z}_\mathcal{J}\|_2$
\end{algorithmic}
\end{algorithm}

Algorithms \ref{alg:embed} and \ref{alg:compare} outline iDist, a fast method for detecting near-duplicate protein--protein interfaces. Algorithm \ref{alg:embed} details the conversion of a protein--protein interface $\mathcal{I}$ into a vector representation $\mathbf{z}_{\mathcal{I}}$. In the first step (line 4), the features of the interface $\mathbf{X}, \mathbf{F}$ and $\mathbf{p}$ are extracted. The residue coordinates $\mathbf{X}$ are determined by the positions of the $C_{\alpha}$ atoms. Next, the residue vector features $\mathbf{F}$ are initialized with simple 20-dimensional one-hot encodings of amino acids. We have also experimented with ESM-1b features \citep{rives2021biological} but obtained slightly lower performance. We observe that the reduced performance could be attributed to ESM-1b biasing the comparison towards entire protein chains rather than the interfaces. Finally, each residue is associated with a label indicating the chain it belongs to, forming the vector $\mathbf{p}$.

\input{assets/fig_idist}

In the following steps (lines 5-12), depicted in \Cref{fig:idist}, the hidden representation $\mathbf{h}_i$ for each residue $i$ is constructed. A residue $i$ receives messages from all other residues in both the same chain ($J_{intra}$) and the partnering chains ($J_{inter}$), represented by exponential radial basis functions with $\alpha$ set to 16. The messages are then averaged to create contact patterns $\mathbf{m}_{intra}$ and $\mathbf{m}_{inter}$. The final representation $\mathbf{h}_i$ is obtained by averaging the difference $\mathbf{m}_{intra} - \mathbf{m}_{inter}$, followed by averaging with the features $\mathbf{f}_i$. The difference heuristic is inspired by the complementarity nature of many non-covalent bonds governing PPIs. When amino acids in both the intra- and inter-contexts of a residue are similar (i.e.~not small--bulky or negatively--positively charged), the message $\mathbf{m}_{intra} - \mathbf{m}_{inter}$ is~low.

Lastly, in steps on lines 13-18, the interface representation $\mathbf{z}_{\mathcal{I}}$ is derived by averaging the hidden features across the individual chains and then across the interaction. As described in Algorithm \ref{alg:compare}, iDist then simply calculates the Euclidean distance between two representations $\mathbf{z}_{\mathcal{I}}$ and $\mathbf{z}_{\mathcal{J}}$ to compare two interfaces $\mathcal{I}$ and $\mathcal{J}$. The outlined procedure can be seen as an implicit approximation of the structural alignment as the resulting representations are $SE(3)$-invariant.

\paragraph{iDist evaluation and threshold selection.} 

\begin{figure}[h!]
  \centering
  \includegraphics[width=0.9\textwidth]{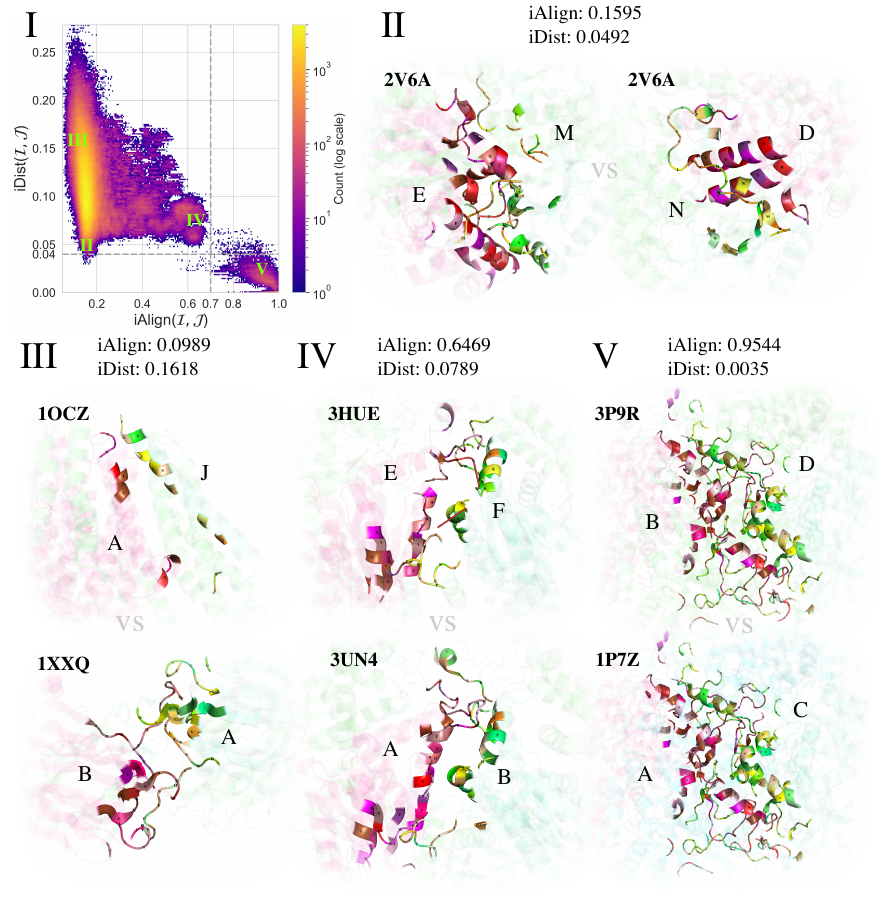}
  \caption[Benchmarking our efficient iDist approximation of the structural alignment algorithm iAlign.]{{\bf Benchmarking our efficient iDist approximation of the structural alignment algorithm iAlign.} (I)~Joint log-scale histogram displaying pair-wise iAlign (IS-score mode, $6\angstrom$ cutoff) and iDist values of 1646 PPI interfaces (2,709,316 pairs) corresponding to 100 PDB codes sampled from DIPS \citep{townshend2019end}. The iAlign values vary between 0 and 1, with high values corresponding to well-aligned interfaces (1 for identical interfaces) and low values corresponding to poorly-aligned interfaces. The iDist varies between 0 to infinity with high values corresponding to structurally-distant interfaces and low values corresponding to similar interfaces (0 for identical interfaces). Figures (III, IV, V) depict samples from regions in (I) where the two methods agree, while (II) shows an example of a disagreement. Each figure displays two interfaces colored by amino acid types, with one protein's palette in reddish hues and the other one in greenish hues. (II) Example of disagreement. The score of iAlign corresponds to the expected value of the alignment of two random PPIs, while iDist suggests higher similarity due to the identity of several fragments of chains M and N (note the $\varepsilon$-like green loop and its further continuation) and similar compositions of amino acids in the helices belonging to proteins E and D (similar combination of reddish colors). In fact, the two interfaces represent different interaction modes of the same two chains in a large symmetric complex. (III) Unrelated interfaces. (IV) Interfaces on the edge of being considered near duplicates. The interactions are obviously related, but the geometry and primary structure differ at every local fragment. (V) Near duplicates. The proteins are visualized using PyMOL \citep{delano2002pymol}.}
  \label{fig:idist_vs_align}
\end{figure}

In order to evaluate the approximation performance of iDist, we compare it with the structural alignmnet algorithm iAlign \citep{gao2010ialign}, the adaptation of TM-score \citep{zhang2004scoring} to protein--protein interfaces. \Cref{fig:idist_vs_align} shows two main modes of the joint probability distribution of the scores output by both methods. For the scores where iAlign($\mathcal{I}, \mathcal{J}) < 0.7$ and iDist$(\mathcal{I}, \mathcal{J}) > 0.04$, the interfaces $\mathcal{I}$ and $\mathcal{J}$ are typically distinct with rare cases of sharing similar structural patterns. In contrast, for the scores where iAlign($\mathcal{I}, \mathcal{J}) > 0.7$ and iDist$(\mathcal{I}, \mathcal{J}) < 0.04$, the interfaces are very likely near duplicates. Notably, for this threshold of 0.04, iDist achieves 0.99\% precision and 0.97\% recall in detecting near-duplicate PPIs with iAlign score greater than 0.7. This observation suggests using the threshold of 0.04 for detecting near-duplicate PPIs with iDist.

To confirm the accurate detection of near-duplicate PPIs with iDist, we additionaly compare our method against independent USalign \citep{zhang2022us} using the same benchmark employed to compare against iAlign. For this, we estimate the similarity score USalign$(\mathcal{I}, \mathcal{J})$ by averaging two output TM-scores corresponding to querying PPI $\mathcal{I}$ against $\mathcal{J}$ and, vice versa, $\mathcal{J}$ against~$\mathcal{I}$ with USalign. As a result of comparison, iDist under the same aforementioned threshold of 0.04 achieves 0.99\% precision and 0.97\% recall with respect to USalign under the threshold of 0.8 (\Cref{fig:idist_vs_usalign}). This result agrees with the comparison of iDist against iAlign. Please note that the optimal near-duplicate thresholds of iAlign (0.7) and USalign (0.8) differ since the methods modify TM-score using different constants.

Based on the described evaluation, we use iDist with the threshold of 0.04 for detecting near-duplicate PPI interfaces that are defined by the distance of $6\angstrom$ between heavy atoms, in our analysis of DIPS composition. Nevertheless, we re-calibrate the threshold to 0.03 for the interfaces defined based on the $10\angstrom$ cutoff distance. This includes the construction of PPIRef50K, used for \ppiformer pre-training.

\paragraph{Construction of PPIRef300K.} Before applying iDist to deduplicate the protein--protein interactions mined from PDB (to construct PPIRef50K), we filter them to only preserve the proper interactions. Specifically, in order to extract only the biochemically proper protein--protein interactions from the raw PPIRef800K dataset and create PPIRef300K, we apply three standard filtering criteria \citep{townshend2019end}. An interaction passes the criteria if: (i) the structure determination method is ``x-ray diffraction'' or ``electron microscopy'', (ii) the crystallographical resolution is $3.5\angstrom$ or better, and (iii) the buried surface area (BSA) of the interface is at least $500{\angstrom}^2$. As a result, 99\% of putative interactions satisfy the method criterion, 79\% the resolution, and 47\% the BSA. We calculate BSA using the \texttt{dr\_sasa} software by \cite{ribeiro2019calculation}.

\section{Implementation details}\label{appendix-implementation}

We implement \ppiformer in PyTorch \citep{paszke2019pytorch} leveraging PyTorch Geometric \citep{fey2019fast}, PyTorch Lightning \citep{Falcon_PyTorch_Lightning_2019}, Graphein \citep{jamasb2020graphein}, and a publicly available implementation of Equiformer\footnote{\url{https://github.com/lucidrains/equiformer-pytorch}}. We pre-train our model on four AMD MI250X GPUs (8 PyTorch devices) in a distributed data parallel (DDP) mode. Our best model was trained for 32 epochs of dynamic masking in 22 hours.

We pre-train \ppiformer by sampling mini-batches of protein--protein interfaces using the Adam optimizer with the default $\beta_1=0.9$, $\beta_2=0.999$ \citep{kingma2014adam}. We partially explore the grid of hyper-parameters given by \Cref{tab:hparams}, and select the best model according to the performance on zero-shot $\Delta \Delta G$ inference on the training set of SKEMPI v2.0. We further fine-tune the model on the same data with the learning rate of $3\cdot10^{-4}$ and sampling 32 mutations per GPU in a single training step, such that each mutation is from a different PPI. We employ the three-fold cross-validation setup discussed in the next section, and ensemble three corresponding fine-tuned models for test predictions. We observe that dividing $C_\alpha$ coordinates by 4, as proposed by \cite{watson2023denovo}, increases the rate of convergence.

\begin{table}[!ht]
\ttabbox[\textwidth]
{
    \caption{Setting of the key pre-training hyper-parameters. The configuration of the best model, referred to as \ppiformern, is highlighted in bold.}
}
{
    \label{tab:hparams}
    \centering
   
    \captionsetup{width=\textwidth}
    \resizebox{\textwidth}{!}{ \renewcommand{\arraystretch}{1}
    \small

    \begin{tabular}{l|l|c}
    \toprule
    Category & Hyper-parameter & Values\\
    \midrule
    \multirow{6}{*}{\shortstack[l]{Data}} & \multirow{2}{*}{\shortstack[l]{Dataset}} & \textbf{PPIRef50K}, PPIRef300K, PPIRef800K, \\
     & & DIPS, iDist-deduplicated DIPS \\
     & Num. neighbors $k$ & \textbf{10}, 30 \\
     & & $\mathbf{\{C_\alpha \rightarrow C_{\beta}^{virtual}\}}$, \\
     & Type-1 features $\mathbf{F}_1$ & $\{C_\alpha \rightarrow C_{\beta}^{virtual}, C_\alpha \rightarrow N, C_\alpha \rightarrow C\}$, \\
     & & $\{C_\alpha \rightarrow C_{\beta}^{virtual}, C_\alpha \rightarrow C_{\alpha}^{prev}, C_\alpha \rightarrow C_{\alpha}^{next}\}$ \\
    \midrule
    \multirow{3}{*}{\shortstack[l]{Masking}} & Fraction & $1, 15\%, 30\%$, $\mathbf{50\%}$ \\
     & Same chain & \textbf{True}, False \\
     & $80\%10\%10\%$ BERT masking & \textbf{True}, False \\
    \midrule
    \multirow{4}{*}{\shortstack[l]{Equiformer}} & Num. layers $L$ & 2, 4, \textbf{8} \\
     & Num. heads $K$ & \textbf{2}, 4, 8 \\
     & Hidden degree $deg$ & \textbf{1} \\
     & Hidden dimensions $\mathbf{d}$ & $\mathbf{[128, 64]}$ \\
    \midrule
    \multirow{2}{*}{\shortstack[l]{Loss}} & Label smoothing $\epsilon$ & $0, \mathbf{0.1}, 0.5$ \\
     & Class weights $\mathbf{w}$ & \textbf{True}, False \\
    \midrule
    \multirow{2}{*}{\shortstack[l]{Optimization}} & Learning rate & $1\cdot10^{-3}, \mathbf{5\cdot10^{-4}}, 3\cdot10^{-4}, 1\cdot10^{-4}$ \\
     & Batch size per GPU & 8, 16, \textbf{32}, 64 \\
    \bottomrule
    \end{tabular}
}
}
\end{table}

\section{Experimental setup}\label{appendix-experimental-details}

This section provides additional details about the experiments with \ppiformern. We describe the choice and setup of the datasets (\Cref{sec:eval-data}), metrics (\Cref{sec:eval-metrics}) and baseline models (\Cref{sec:baselines}).

\subsection{Test datasets}\label{sec:eval-data}

\paragraph{SKEMPI v2.0.}
SKEMPI v2.0 \citep{jankauskaite2019skempi} is an invaluable resource for training and evaluating \ddg predictors. Since its initial release \citep{moal2012skempi}, the dataset has played a pivotal role in advancing the field of computational PPI design \citep{geng2019finding}. However, despite encompassing mutations from nearly 300 available studies, it still only contains 7085 mutations across only 343 PPI structures. Moreover, the dataset is highly biased, with nearly three-quarters of the data corresponding to single-point mutations, more than half of which are mutations to alanine. Additionally, the dataset contains biases towards specific types of PPIs, including the repetition of near-duplicate structures or identical mutations with slightly different \ddg annotations. While it is difficult to enhance the quality or diversity of the dataset without additional wet lab experiments, it is crucial for the machine learning research community to leverage the dataset for training models that generalize beyond the available data. Here, we reconsider the data splitting strategy employed for training and evaluating state-of-the-art models and propose a new scheme based on an in-depth analysis of the original publication by \cite{jankauskaite2019skempi}.

In contrast to previous works \citep{luo2023rotamer, shan2022deep, liu2021deep}, we do not use a PDB-disjoint split of SKEMPI v2.0 (based on the \texttt{\#Pdb} column in the dataset) for training and assessing the performance of \ddg predictors. We find that this approach leads to at least two kinds of data leakage, potentially limiting the generalization of models towards unseen binders. To illustrate this leakage, we analyze a recent cross-validation split used to train and test the RDE-Network model \citep{luo2023rotamer}.

First, we observe that every third test mutation in the split is, in fact, a mutation of the training PPI with a different PDB code in the test set (but with the same values in the \texttt{Protein~1} and \texttt{Protein~2} columns). For instance, PPIs with the \texttt{2WPT\_A\_B} and \texttt{1EMV\_A\_B} codes have nearly identical 3D structures, representing the same interaction between the colicin toxin and an immune protein of \textit{E.~coli}. These structures have 64 and 59 annotated mutations, respectively, with 21 mutations common to both structures. However, due to distinct \texttt{\#Pdb} codes, \texttt{2WPT\_A\_B} and \texttt{1EMV\_A\_B} end up in different folds in \citep{luo2023rotamer}, resulting in data leakage on the level of mutations.

Second, over half of the test \texttt{\#Pdb} codes violate the original hold-out separation proposed by the authors of SKEMPI v2.0. Specifically, \cite{jankauskaite2019skempi} performed automated clustering of PPIs followed by manual inspection to split entries into independent groups to assess machine learning generalization, resulting in the \texttt{Hold\_out\_proteins} column in the dataset. We observe that 57\% of test PPIs in the PDB-disjoint split in \citep{luo2023rotamer} violate the proposed grouping, having a homologous PPI with the same \texttt{Hold\_out\_proteins} value in the training set. For example, the first two PPIs in the SKEMPI v2.0 table, \texttt{1CSE\_E\_I} and \texttt{1ACB\_E\_I}, are both interactions between a serine protease and the protein inhibitor Eglin c. These PPIs, coming from the same study \citep{qasim1997interscaffolding}, have 6 identical annotated mutations of Eglin c with similar \ddg values. As a result, \cite{jankauskaite2019skempi} assigned these PPIs to the same \texttt{Hold\_out\_proteins} group, \texttt{Pr/PI}, representing protease--protein inhibitor interactions. Nevertheless, since the PPIs have different PDB codes (\texttt{\#Pdb}), they got separated into different folds in \citep{luo2023rotamer}, which demonstrates the data leakage on the level of proteins.

Therefore, to ensure effective evaluation of generalization and mitigate the risk of overfitting, we divide SKEMPI v2.0 into 3 cross-validation folds based on the \texttt{Hold\_out\_proteins} feature, as originally proposed by the dataset authors. Additionally, we stratify the $\ddg$ distribution across the folds to ensure balanced labels. Before constructing the cross-validation split, we reserve 5 distinct PPIs to create 5 test folds. We consider a PPI distinct if it does not share interacting partners or homologous binding site \citep{jankauskaite2019skempi} with any other PPI in SKEMPI v2.0\footnote{Figure 2 in the original SKEMPI v2.0 paper \citep{jankauskaite2019skempi} illustrates such distinct PPIs by connected components consisting of two nodes (proteins), or more than two nodes but with two unique ones, where uniqueness is defined by the ``Share common binding site'' edges.} (and, consequently, has a unique \texttt{Hold\_out\_proteins} value). Furthermore, we ensure that for each of the 5 selected interactions, both negative and positive \ddg labels are present. Finally, the maximum iAlign IS-score between all PPIs in the test folds and those in the train-validation folds is 0.22, confirming the intended structural independence of the test set. Please refer to \Cref{fig:skempi-test} for details on the constructed test set.

\paragraph{Optimization of a human antibody against SARS-CoV-2.}

We utilize the benchmark of \cite{luo2023rotamer} to test the capabilities of models to optimize a human antibody against \mbox{SARS-CoV-2} \citep{shan2022deep}. Specifically, the goal is to retrieve 5 mutations on the heavy chain CDR region of the antibody that are known to enhance the neutralization effectiveness within a pool of all exhaustive 494 single-point mutations on the interface. Please note that the effects of the other 489 mutations are considered unknown and may be either favorable or unfavorable.

\paragraph{Engineering staphylokinase for enhanced thrombolytic activity.} 

We assess the potential of the methods to enhance the thrombolytic activity of the staphylokinase protein (SAK). Staphylokinase, known for its cost-effectiveness and safety as a thrombolytic agent, faces a significant limitation in its widespread clinical application due to its low affinity to plasmin \citep{nikitin2022computer}. In this study, we leverage 80 thrombolytic activity labels associated with SAK mutations located at the binding interface with plasmin \citep{laroche2000recombinant}. 

Specifically, we evaluate the $\Delta \Delta G$ predictions on the 1BUI structure from PDB, which contains the trimer consisting of plasmin-activated staphylokinase bound to another plasmin substrate. Unlike in the case of the SARS-CoV-2 benchmark, all 80 binary labels for SAK--plasmin have experimentally measured effects, among which 20 mutations are favorable. Additionally, 6 of them introduce at least two-fold thrombolytic activity improvement, constituting even more practically-significant targets for retrieval. Besides that, 24 out of 80 mutations on SAK are multi-point, which provides a more general setup for the evaluation than the SARS-CoV-2 benchmark. Note that while the quantity measured for SAK is the activity, what we are estimating is the \ddg for the SAK--plasmin interaction. Since the affinity of the complex is the main activity bottleneck, these two quantities were shown to be highly correlated \citep{nikitin2022computer}.

\subsection{Evaluation metrics}\label{sec:eval-metrics}

In a practical binder design scenario \citep{nikitin2022computer, shan2022deep}, the primary objectives typically include (i) prioritizing mutations with lower \ddg values, and more specifically, (ii) identifying stabilizing mutations (with negative \ddgn) from a pool of candidates. To address~(i), we evaluate models using the Spearman correlation coefficient between the ground-truth and predicted \ddg values. For (ii), we calculate precision and recall with respect to mutations with negative \ddgn. We calculate metrics for each protein--protein interaction separately, ensuring that mutations from different interactions are not combined. Subsequently, we average these results to obtain aggregated metric values. This approach estimates the expected performance of a model on a new, unseen PPI.

To enable comparisons with other methods, we also report metrics used in previous works: Pearson correlation coefficient, mean absolute error (MAE) and root mean squared error (RMSE), as well as area under the receiver operating characteristic (ROC AUC) with respect to the sign of mutations. However, we stress that these metrics may be misleading when selecting a model for a practical application. Pearson correlation may not capture the non-linear scoring capabilities of a method and is sensitive to outlying and less significant predictions on destabilizing mutations. MAE and RMSE are not invariant to monotonic transformations of predictions, and ROC AUC may overemphasize the performance on destabilizing mutations. 

On the independent SARS-CoV-2 and SAK engineering case studies, we report scores for each of the favorable mutations following \cite{luo2023rotamer}. In addition, we evaluate more systematic precision at 1 (P@1) and precision at $k\%$ on the total pool of mutations (P@$k\%$) for $k \in \{5, 10\}$.

\subsection{Baselines}\label{sec:baselines}

\paragraph{Flex ddG {\normalfont \citep{barlow2018flex}}.}
Flex ddG is the most advanced Rosetta \citep{leman2020rosetta} protocol which predicts \ddg by estimating the change in binding free energies between the wild type and the mutant structures using force field simulations. The same protocol is used in a recent RosettaDDGPrediction toolbox \citep{sora2023rosettaddg}. For each mutation, the \ddg prediction is obtained by averaging the output from 5 runs of \texttt{ddG-no\_backrub\_control}, a number shown to be optimal by \cite{barlow2018flex}. Since on average the prediction of a single \ddg on the SKEMPI v2.0 test folds using flex ddG requires approximately 1 CPU hour (on Intel Xeon Gold 6130), we do not evaluate the complete \texttt{ddG-backrub} protocol. When using the \texttt{ddG-backrub} parameters suggested by \cite{barlow2018flex}, the running time further increases by orders of magnitude, making the method impractical when compared to other baseline methods. For comparison, on the same data where flex ddG takes on average 1 CPU hour per mutation, our \ppiformer requires 73 GPU milliseconds per mutation (using one out of two devices on AMD MI250X GPU with the batch size of 1). 

\paragraph{GEMME {\normalfont \citep{laine2019gemme}}.} Global Epistatic Model for predicting Mutational Effects (GEMME) is a state-of-the-art sequence-based predictor that does not involve machine learning. The method explicitly models the evolutionary history of a sequence and derives scores of variants based on the evolutionary conservation of amino acids. GEMME is currently the second-best method in the ProteinGym benchmark\footnote{\url{https://proteingym.org/benchmarks}} which was created to assess the performance of models in predicting protein fitness from deep mutations scanning data \citep{notin2022tranception}.

\paragraph{MSA Transformer {\normalfont \citep{rao2021msa}}.}
MSA Transformer is an unsupervised sequence-based baseline, trained in a self-supervised way on diverse multiple sequence alignments (MSA). We use the pre-trained model provided by \cite{rao2021msa} and build MSAs against UniRef30 database using HHblits algorithm \citep{Remmert2012} with the same parameters as were used to train the MSA Transformer. MSA Transformer can be applied to score mutations in a few-shot setup relying on masked pre-training \citep{meier2021language}. More specifically, the $\Delta \Delta G$ can be predicted as a difference of the log-likelihoods of the wild-type and mutant sequences conditioned on a provided~MSA.

\paragraph{ESM-IF {\normalfont \citep{hsu2022learning}}.}

ESM-IF, an inverse folding model based in GVP-GNN \citep{jing2020learning}, has been demonstrated by \cite{hsu2022learning} to effectively generalize for predicting the signs of \ddg values for single-point mutations in SKEMPI \citep{jankauskaite2019skempi}. To predict $\Delta G$, the authors compute the average log-likelihood of amino acids within a mutated chain, conditioned on the backbone structure of the complex. As done by the authors, to predict \ddgn, we subtract the likelihood of a mutant chain from that of the wild-type chain. Additionally, to account for simultaneous mutations across multiple chains, we perform an ESM-IF forward pass for each chain in a complex and average their likelihoods to obtain the final result. Our evaluation of ESM-IF on SKEMPI v2.0 leads to results consistent with the original publication. Specifically, ESM-IF achieves high ROC AUC of 0.68 on classifying the sign of mutations (\Cref{fig:skempi_test}), similarly to the ROC AUC of 0.71 reported for single-point mutations in \citep[Table~C.5]{hsu2022learning}.

\paragraph{RDE-Network {\normalfont \citep{luo2023rotamer}}.}

To validate the performance of RDE-Network on our 5 test folds of SKEMPI v2.0, we exclude all test examples from cross-validation and retrain the model with the same hyperparameters as used by \cite{luo2023rotamer}. Since the exclusion of data points affects the sampling of training batches, and may, therefore, negatively affect the performance, we use 3 different random seeds and choose the final model according to the minimal validation loss. For the evaluation on SKEMPI-independent case studies, we use pre-trained model weights as provided by \cite{luo2023rotamer}.

\section{Ablations}\label{sec:appendix-ablations}

This section evaluates the effects of the key components proposed to enhance the generalization capabilities of \ppiformern. We first demonstrate the importance of using the PPIRef50K dataset for pre-training, as well as employing the proposed pre-training regularization techniques (\Cref{sec:ablations-ssl}). Next, in  \Cref{sec:ablations-ddg} we demonstrate the positive impact of pre-training on the subsequent \ddg prediction task, and highlight the importance of the log odds ratio. Additionally, we verify the stability of the fine-tuning process across different random seeds.

\subsection{Self-supervised pre-training}\label{sec:ablations-ssl}

\input{assets/fig_ablations}

\Cref{fig:ablations}A (Top) illustrates that \ppiformer achieves a notable performance without any supervised fine-tuning (per-PPI $\rho_{Spearman} = 0.21$ and precision~=~35\% on 5,643 mutations from SKEMPI v2.0). Next, we discuss the importance of the choice of the pre-training dataset used, as well as the proposed regularization measures.

\paragraph{Pre-training dataset.}

First of all, the best pre-training is achieved when sampling protein--protein interactions from our PPIRef50K dataset, both on scoring mutations and detecting the stabilizing ones (a3). We observe that training from redundant PPIRef300K decreases the performance, most probably because of introducing biases towards over-represented proteins and interactions into the model (a4). Finally, training from raw putative PPIs (a5) achieves the worst performance while training from DIPS or DIPS-Plus (a1, a2)  is still worse than our PPIRef~(a3).

\paragraph{Pre-training regularization.}

Next, we evaluate the benefits of three key components of \ppiformern: masking strategy, data representation and the loss function. First, the $80\%10\%10\%$ masking (i.e.~replacing 10\% of masked nodes with native amino acids and 10\% with random ones \citep{devlin2018bert}) leads to better performance (\ppiformer vs. b1), as well as masking only residues from the same chain (\ppiformer vs. b2), which better corresponds to practical binder design scenarios. Next, as discussed in \Cref{subsec:representation}, we represent the  orientations of residues with a single virtual beta-carbon vector per residue. This is in contrast with the more widely adopted utilization of three vectors per residue describing the complete frame of an amino acid by including the direction to a real $C_{\beta}$ atom and the directions to alpha-carbons of neighboring residues \citep{jing2020learning} or, alternatively, two orthonormalized vectors in the directions of neighboring $N$ and $C$ atoms along the protein backbone \citep{watson2023denovo, yim2023se}. We observe that a single virtual beta carbon (used in our \ppiformern) leads to better generalization compared to extra full-frame representations (c1, c2). Finally, the regularization by label smoothing and amino acid class weighting have additional strong positive impact on the generalization capabilities of \ppiformern. Switching off the regularization measures significantly lowers the performance~(d1-d3).

\subsection{Fine-tuning for $\Delta \Delta G$ prediction}\label{sec:ablations-ddg}

\paragraph{Effect of pre-training on fine-tuning.}

We further ablate the importance of our self-supervised pre-training for the $\Delta \Delta G$ fine-tuning. \Cref{fig:ablations}B illustrates the crucial significance of the pre-training. The pre-trained \ppiformer surpasses the randomly initialized model with a margin of 0.08 absolute difference on per-PPI Spearman correlation. Note that the pre-trained zero-shot predictor (red curve at step 0) achieves performance competitive to the fully trained model without pre-training (green curve, highest peak).

\paragraph{Fine-tuning head.}

In \Cref{subsec:fine-tuning}, we discuss the motivation of employing the log odds ratio for \ddg fine-tuning. The choice of log odds is strongly justified within the context of binding thermodynamics, offering the advantageous inductive bias of antisymmetry. Furthermore, this approach is compatible with the log-likelihood-based masking strategy employed during pre-training, ensuring a more effective transfer learning. We validate our argument by performing two ablation experiments.

First, following \cite{liu2021deep}, we implement a baseline regressor that performs two forward passes to predict $\Delta \Delta G$: one for the wild-type structure and another for a mutated structure (i.e., with replaced  $\mathbf{F}_0$ features capturing one-hot encoded amino acid types). Then, it estimates $\Delta \Delta G$ by taking the average of node embeddings corresponding to residues being mutated in both structures, concatenating them, and applying a simple two-layer regressor of the form: $Linear(128, 64) \rightarrow ReLU \rightarrow Linear(64, 1)$. This approach results in a significant drop in performance (see the first row ``Naive regressor'' in \Cref{tab:fine_tuning_kind}).

Second, we improve this naive design by incorporating antisymmetry: removing biases in the $Linear$ layers, replacing $ReLU$ with $Tanh$ (an odd non-linear function), and subtracting averaged embeddings instead of concatenating them. This improvement leads to a performance boost compared to the naive baseline, but still strongly underperforms our log odds ratio (\ppiformern) in terms of practically important metrics (see the second row ``Antisymmetric regressor'' in \Cref{tab:fine_tuning_kind}).

\input{assets/table_fine_tuning_kind}

\paragraph{Fine-tuning stability.}

Finally, we demonstrate that the test results are consistent under different random seeds used for fine-tuning (\Cref{tab:fine_tuning_seed}). Consequently, we report the average over the three different seeds and the standard deviation in other tables.

\input{assets/table_fine_tuning_seed}

\section{Additional results}\label{appendix-additional-results}

\paragraph{SKEMPI v2.0 test set.}

\Cref{fig:skempi_test} shows the average performance of all compared methods on 5 test folds from SKEMPI v2.0. In this section, we provide a more detailed comparison. Specifically, \Cref{fig:skempi-test} demonstrates non-aggregated performance of the methods on all test PPIs. Notably, \ppiformer achieves state-of-the-art performance by attaining the best and second-best metric values in 7 and 4 cases, respectively, compared to the best competing methods GEMME (6 and 1) and RDE-Network (5 and 2).

Please note that in some cases methods fail to make \ddg predictions. For example, MSA Transformer fails to predict the effects of 4 mutations on the “dHP1 Chromodomain and H3 tail” interaction. This is caused by HHblits returing an empty MSA for the short binding peptide of 8 amino acids. To ensure a fair comparison, in such cases we impute predictions with the average \ddg value from the SKEMPI v2.0 training set, which is approximately $0.69$.

\input{assets/table_fig_SKEMPI2_test}

\input{assets/fig_idist_usalign}

\end{document}

%% file: math_commands.tex
\usepackage{amsmath,amsfonts,bm}

\def\eqref#1{equation~\ref{#1}}
\def\1{\bm{1}}

\DeclareMathAlphabet{\mathsfit}{\encodingdefault}{\sfdefault}{m}{sl}
\SetMathAlphabet{\mathsfit}{bold}{\encodingdefault}{\sfdefault}{bx}{n}

%% file: assets/datasets_table.tex
\begin{wrapfigure}{R}{6.5cm}

\vspace{-0.3cm}

\ttabbox[\textwidth]
{
    \caption{Comparison of PPIRef with existing datasets of native protein complexes.}
}
{
    \setcounter{table}{0}
    \label{tab:ppi_data_stats}
    \centering   
    \captionsetup{width=\textwidth}
    \caption{Comparison of our PPIRef dataset with existing datasets of native protein complexes. The number of unique interfaces is estimated by deduplication using our iDist algorithm.}
    \resizebox{\textwidth}{!}{ \renewcommand{\arraystretch}{1.1}
    \small
    \begin{tabular}{lrrr}
    \toprule
    \multirow{2}{*}{Dataset} & PPI & Unique \\ & structures & interfaces \\
    \midrule
    MaSIF-search & 6K & 5K \\
    DIPS / DIPS-Plus & 40K  & 9K \\
    PPIRef (ours) & \textbf{322K} & \textbf{46K} (PPIRef50K) \\
    \bottomrule
    \end{tabular}
}
}

\vspace{-0.2cm}

\end{wrapfigure}

%% file: assets/duplicates_example.tex
\begin{wrapfigure}{R}{6.5cm}

\vspace{-0.45cm}

\ffigbox[\textwidth]
{
    \caption{An example of protein--protein interfaces from different folds of the DIPS dataset detected as near duplicates by our iDist method. Both PPIs come from the same KatE enzyme homooligomers with different single-point mutations. Notably, the symmetry of the complex itself yields 3 groups of 2 duplicates (from a single PDB entry with 6 PPIs). Furthermore, querying PDB with ``KatE'' results in 36 KatE complexes, yielding, therefore, 3 groups of 72 duplicates each. Our iDist approach can efficiently identify such structural near duplicates on a large scale.
    }
}
{
    \label{fig:dips_duplicates}
    \includegraphics{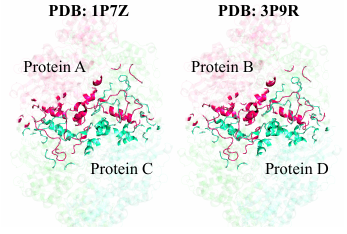}
}

\vspace{-0.5cm}

\end{wrapfigure}

%% file: assets/table_skempi_test_overall.tex
\begin{table}[!ht]
\ttabbox[\textwidth]
{
    \caption{Test set performance averaged across five held-out protein--protein interactions selected from SKEMPI v2.0 for benchmarking (see \Cref{sec:eval-data} for details). The standard deviation for \textsc{PPIformer} is estimated from three fine-tuning experiments with different random seeds.}
    \vspace*{-2mm}
}
{
\resizebox{\textwidth}{!}{\renewcommand{\arraystretch}{1.1}
\small
\label{fig:skempi_test}
\begin{tabular}{ll|ccccccc}
\toprule
Category & Method &  Spearman $\uparrow$ &   Pearson $\uparrow$ &  Precision $\uparrow$ &    Recall $\uparrow$ &   ROC AUC $\uparrow$ &       MAE $\downarrow$ &      RMSE $\downarrow$ \\
\midrule
\multirow{2}{*}{\shortstack[l]{Force field \\ simulations}} & \multirow{2}{*}{\textsc{Flex ddG}} & \multirow{2}{*}{$0.55$} & \multirow{2}{*}{$0.57$} & \multirow{2}{*}{$0.63$} & \multirow{2}{*}{$0.62$} & \multirow{2}{*}{$0.84$} & \multirow{2}{*}{$1.59$} & \multirow{2}{*}{$2.00$} \\
 & & & & & & & & \\
\midrule
\multirow{5}{*}{\shortstack[l]{Machine \\ learning}} 
& \textsc{GEMME}                   &  \underline{0.38} &   \underline{0.41} &      $\mathbf{0.60}$ &  0.49 &  \underline{0.74} &  2.16 &  2.81 \\
& \textsc{MSA Transformer} & $0.31$ & $0.36$ & $0.51$ & $0.38$ & $0.70$ & $6.13$ & $6.93$ \\
& \textsc{ESM-IF} & $0.18$ & $0.18$ & $0.33$ & $0.41$ & $0.68$ & $1.87$ & $2.15$ \\
& \textsc{RDE-Net.} & $0.24$ & $0.30$ & $\underline{0.54}$ & $\mathbf{0.65}$ & $0.67$ & $\underline{1.70}$ & $\underline{2.02}$ \\
& \textsc{PPIformer (ours)} & $\mathbf{0.44} \pm 0.03$           & $\mathbf{0.46} \pm 0.03$         & $\mathbf{0.60} \pm 0.014$         & $\underline{0.61} \pm 0.012$     & $\mathbf{0.78} \pm 0.019$     & $\mathbf{1.64} \pm 0.011$    & $\mathbf{1.94} \pm 0.006$ \\
\bottomrule
\end{tabular}
}
}
\end{table}

%% file: assets/table_ranking_covid.tex
\begin{table}[!ht]
\ttabbox[\textwidth]
{
    \caption{The performance of \ppiformer and eight competing methods in retrieving 5 human antibody mutations effective against SARS-CoV-2. The middle section shows ranks (in percent) for individual mutations with predictions scored in the top 10\% in bold, as in \citep{luo2023rotamer}. The right section presents precision metrics. The values for baseline methods except for MSA Transformer and ESM-IF are reproduced from \citep{luo2023rotamer}.}
    \vspace*{-2mm}
}
{
\resizebox{\textwidth}{!}{ \renewcommand{\arraystretch}{1.1}
\small
\label{tab:res-covid}
\begin{tabular}{l|ccccc|ccc}
    \toprule
    Method & \texttt{TH31W} $\downarrow$ & \texttt{AH53F} $\downarrow$ & \texttt{NH57L} $\downarrow$ & \texttt{RH103M} $\downarrow$ & \texttt{LH104F} $\downarrow$ & P@1 $\uparrow$ & P@5\% $\uparrow$ & P@10\% $\uparrow$\\
    \midrule
    \textsc{MSA Transformer} & 56.88 & 42.11 & 63.56 & 49.19 & 18.83 & \underline{0.00} & 0.00 & 0.00 \\
\textsc{ESM-IF} & 49.39 & 17.61 & 17.00 & 51.42 & 48.58 & \underline{0.00} & 0.00 & 0.00 \\
\textsc{Rosetta} & 10.73 & 76.72 & 93.93 & 11.34 & 27.94 & \underline{0.00} & 0.00 & 0.00 \\
\textsc{FoldX} & 13.56 & \textbf{6.88} & \textbf{5.67} & 16.60 & 66.19 & \underline{0.00} & 0.00 & 4.08 \\
\textsc{DDGPred} & 68.22 & \textbf{2.63} & 12.35 & \textbf{8.30} & \textbf{8.50} & \underline{0.00} & \underline{4.00} & \textbf{6.12} \\
\textsc{End-to-End} & 29.96 & \textbf{2.02} & 14.17 & 52.43 & 17.21 & \underline{0.00} & \underline{4.00} & 2.04 \\
\textsc{MIF-Net.} & 24.49 & \textbf{4.05} & \textbf{6.48} & 80.36 & 36.23 & \underline{0.00} & \underline{4.00} & \underline{4.08} \\
\textsc{RDE-Net.} & \textbf{1.62} & \textbf{2.02} & 20.65 & 61.54 & \textbf{5.47} & \underline{0.00} & \textbf{8.00} & \textbf{6.12} \\
\textsc{PPIformer (ours)} & 18.02 & \textbf{0.20} & \textbf{7.69} & 21.46 & 10.9 & \textbf{100} & \underline{4.00} & \underline{4.08} \\
    \bottomrule
\end{tabular}
}
}
\end{table}

%% file: assets/table_ranking_sak.tex
\begin{table}[!ht]
\ttabbox[\textwidth]
{
    \caption{The performance of \ppiformer and three competing methods in retrieving the mutations on the staphylokinase interface that increase its trombolytic activity via enhanced affinity to plasmin. Here, 20 of 80 mutations are favorable for activity (right section), and 6 of them lead to at least two-fold enhancement (middle section). Similarly to \Cref{tab:res-covid}, the rank values in bold are those in the top-10\% predictions.}
    \vspace*{-2mm}
}
{
\label{tab:sak}
\resizebox{\textwidth}{!}{ \renewcommand{\arraystretch}{1.1}
\small
\begin{tabular}{l|cccccc|ccc}
\toprule
\multirow{4}{*}{\shortstack[l]{Method}} & \multicolumn{6}{c|}{Mutations with $\geq 2 \times$ activity enhancement} & \multicolumn{3}{c}{\multirow{4}{*}{Activity enhancement}} \\
& {} & {} & {} & {} & \texttt{KC74Q}\hphantom{ $\downarrow$} & \texttt{KC74R}\hphantom{ $\downarrow$} & {} & {} & {} \\
{} & {} & {} & \texttt{KC135R}\hphantom{ $\downarrow$} & {} & \texttt{KC130E}\hphantom{ $\downarrow$} & \texttt{KC130T}\hphantom{ $\downarrow$} & {} & {} & {} \\
{} & \texttt{KC130A} $\downarrow$ & \texttt{KC130T} $\downarrow$ & \texttt{KC130T} $\downarrow$ & \texttt{KC135A} $\downarrow$ & \texttt{KC135R} $\downarrow$ & \texttt{KC135R} $\downarrow$ & P@1 $\uparrow$ &  P@5\% $\uparrow$ &  P@10\% $\uparrow$ \\
\midrule
\textsc{MSA Transformer} & 52.50 & 32.50 & 55.00 & 40.00 & 70.00 & 78.75 & \textbf{100} & \underline{50.00} & 37.50 \\
\textsc{ESM-IF} & 45.00 & 33.75 & 46.25 & 25.00 & 42.50 & 58.75 & \underline{0.00} & 0.00 & 25.00 \\
\textsc{RDE-Net.} & 51.25 & 33.75 & 22.50 & 15.00 & 27.50 & \textbf{5.00} & \underline{0.00} & \underline{50.00} & \underline{62.50} \\
\textsc{PPIformer (ours)} & 66.25 & 15.00 & \textbf{2.50} & 52.50 & 33.75 & \textbf{1.25} & \textbf{100} & \textbf{75.00} & \textbf{87.50} \\
\bottomrule
\end{tabular}
}
}
\vspace*{-3mm}
\end{table}

%% file: assets/fig_idist.tex
\begin{wrapfigure}{R}{4.2cm}

\vspace{-0.4cm}

\ffigbox[\textwidth]
{
    \caption{Illustration of the single-step message passing of iDist (lines 5-12 in \Cref{alg:embed}). A residue $i$ receives complementary distance-weighted messages $\mathbf{m}_{intra}$ and $\mathbf{m}_{inter}$ from all residues within the same protein $J_{intra}$ and other partners $J_{inter}$. The messages are aggregated into the embedding $\mathbf{h}_i$ and the procedure is repeated for each interface residue. iDist efficiently detects near-duplicate PPIs as the ones having similar averaged interface embeddings.
    }
}
{
    \label{fig:idist}
    \includegraphics[scale=0.5]{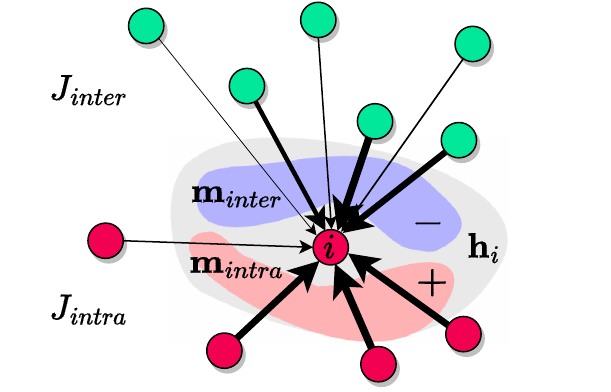}
}

\vspace{-0.9cm}

\end{wrapfigure}

%% file: assets/fig_ablations.tex
\begin{figure}[h!]
  \centering
  \includegraphics[width=1.0\textwidth]{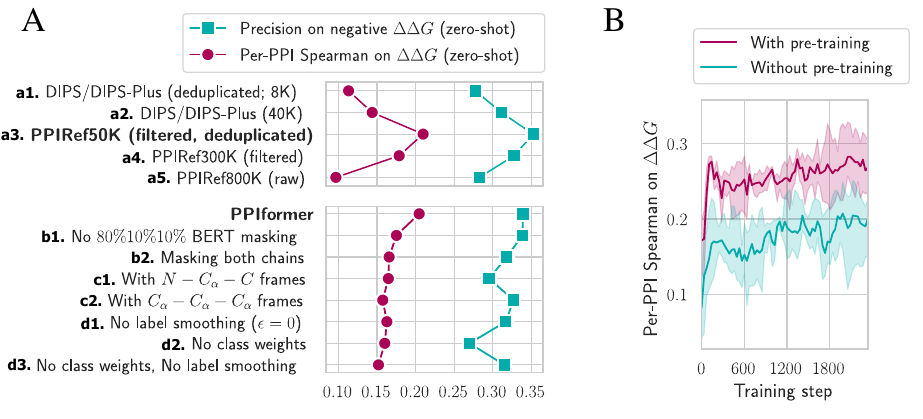}
  \caption[Pre-training and fine-tuning ablations.]{\textbf{Pre-training and fine-tuning ablations}. (A) Pre-training ablations with respect to datasets (a1-a5) and \ppiformer masking strategies (b1-b2), input protein representations (c1-c2), and loss functions (d1-d3). The metrics show zero-shot performance on the training fold of the SKEMPI v2.0 dataset, i.e. $\Delta \Delta G$ inference according to \Cref{eq:ddg-pred} without any supervision. Each row in the figure represents a single modification of the complete \ppiformer model. Dataset ablations precede model ablations, using different SKEMPI v2.0 subsets. The dataset ablations (a1-a5) are done on two of the three folds from \citep{luo2023rotamer}, and the model ablations (b1-b2) use the training set described in \Cref{sec:eval-data}. Thus, the performance varies slightly between (a3) and \ppiformer rows (b1-b2), despite representing the same model. (B) The effect of pre-training on $\Delta \Delta G$ fine-tuning. Each training step is performed by sampling a single mutation from each PPI in a randomly sampled batch. The shaded areas correspond to the minimum, mean and maximum of the performance of three models in three-fold cross-validation.}
  \label{fig:ablations}
\end{figure}

%% file: assets/table_fine_tuning_kind.tex
\begin{table}[!ht]
\ttabbox[\textwidth]
{
    \caption{Performance of \ppiformer with different fine-tuning heads. The standard deviation for \ppiformer is estimated from three fine-tuning experiments with different random seeds.}
}
{
\resizebox{\textwidth}{!}{\renewcommand{\arraystretch}{1.1}
\small
\label{tab:fine_tuning_kind}
\begin{tabular}{l|ccccccc}
\toprule
Method &  Spearman $\uparrow$ &   Pearson $\uparrow$ &  Precision $\uparrow$ &    Recall $\uparrow$ &   ROC AUC $\uparrow$ &       MAE $\downarrow$ &      RMSE $\downarrow$ \\
\midrule
\begin{tabular}[c]{@{}l@{}}\textsc{PPIformer} \\ (Naive regressor)\end{tabular} & $0.298$ & $0.330$    & $0.204$  & $0.258$   & $\underline{0.724}$& $1.844$   & $2.210$ \\
\begin{tabular}[c]{@{}l@{}}\textsc{PPIformer} \\ (Antisymmetric regressor)\end{tabular} & $\underline{0.348}$      & $\underline{0.382}$    & $\underline{0.494}$       & $\textbf{0.626}$   & $0.658$     & $\underline{1.720}$   & $\underline{1.964}$ \\
\begin{tabular}[c]{@{}l@{}}\textsc{PPIformer} \\ (Ours, \Cref{subsec:fine-tuning})\end{tabular} & $\textbf{0.44} \pm 0.03$ & $\textbf{0.46} \pm 0.03$    & $\textbf{0.60} \pm 0.014$  & $\underline{0.61} \pm 0.012$   & $\textbf{0.78} \pm 0.019$ & $\textbf{1.64} \pm 0.011$   & $\textbf{1.94} \pm 0.006$ \\
\bottomrule
\end{tabular}
}
}
\end{table}

%% file: assets/table_fine_tuning_seed.tex
\begin{table}[!ht]
\ttabbox[\textwidth]
{
    \caption{Performance of \ppiformer under different fine-tuning seeds.}
    % \vspace*{-2mm}
}
{
\resizebox{\textwidth}{!}{\renewcommand{\arraystretch}{1.1}
\small
\label{tab:fine_tuning_seed}
\begin{tabular}{l|ccccccc}
\toprule
Method &  Spearman $\uparrow$ &   Pearson $\uparrow$ &  Precision $\uparrow$ &    Recall $\uparrow$ &   ROC AUC $\uparrow$ &       MAE $\downarrow$ &      RMSE $\downarrow$ \\
\midrule
\begin{tabular}[c]{@{}l@{}}\textsc{PPIformer} \\ (Ours)\end{tabular} & $0.44 \pm 0.03$           & $0.46 \pm 0.03$         & $0.60 \pm 0.014$         & $0.61 \pm 0.012$     & $0.78 \pm 0.019$     & $1.64 \pm 0.011$    & $1.94 \pm 0.006$ \\
\begin{tabular}[c]{@{}l@{}}\textsc{PPIformer} \\ (Ours, seed 1)\end{tabular} & $0.42$                     & $0.46$                   & $0.58$                   & $0.61$               & $0.77$               & $1.65$              & $1.94$ \\
\begin{tabular}[c]{@{}l@{}}\textsc{PPIformer} \\ (Ours, seed 2)\end{tabular} & $0.47$                     & $0.49$                   & $0.60$                   & $0.62$               & $0.80$               & $1.62$              & $1.94$ \\ 
\begin{tabular}[c]{@{}l@{}}\textsc{PPIformer} \\ (Ours, seed 3)\end{tabular} & $0.42$                     & $0.43$                   & $0.61$                   & $0.60$               & $0.77$               & $1.64$              & $1.95$ \\
\bottomrule
\end{tabular}
}
}
\end{table}

%% file: assets/table_fig_SKEMPI2_test.tex
\begin{table}
  \centering
    \begin{minipage}{.49\textwidth}
    \resizebox{\textwidth}{!}{ \renewcommand{\arraystretch}{1.1}
    \small
    \begin{tabular}{l|ccc}
        \toprule
        Method & Spearman & Precision & Recall \\
        \midrule
        \textsc{Flex ddG} & 0.68 & 100\% & 50.00\% \\
        \midrule
        \textsc{GEMME} & 0.61 &	\textbf{100\%}	& \textbf{100\%} \\
        \textsc{MSA Transformer} & 0.61 &	\textbf{100\%}	& \textbf{100\%} \\
        \textsc{ESM-IF} & 0.34 & \underline{50.00\%} & \underline{50.00\%} \\
        \textsc{RDE-Net.} & \underline{0.68} & \textbf{100\%} & \textbf{100\%} \\
        \textsc{PPIformer} & \textbf{0.75} & \textbf{100\%} & \textbf{100\%} \\
        \bottomrule
    \end{tabular}
    }
    \small{Complement C3d and Fibrinogen-binding protein Efb-C (9 mutations, 4 neg., 5. pos.)}
    \end{minipage}%
  \hfill
  \begin{minipage}{.49\textwidth}
    \resizebox{\textwidth}{!}{ \renewcommand{\arraystretch}{1.1}
    \small
    \begin{tabular}{l|ccc}
        \toprule
        Method & Spearman & Precision & Recall \\
        \midrule
        \textsc{Flex ddG} & 0.82 & 42.86\% & 42.86\% \\
        \midrule
        \textsc{GEMME} & 0.40	& \textbf{100}\% & \underline{64.29}\% \\
        \textsc{MSA Transformer} & 0.43	& \underline{87.50}\% & 50.00\% \\
        \textsc{ESM-IF} & 0.18 & 41.18\% & 50.00\% \\
        \textsc{RDE-Net.} & \underline{0.58} & 42.11\% & 57.14\% \\
        \textsc{PPIformer} & \textbf{0.60} & 38.46\% & \textbf{71.43\%} \\
        \bottomrule
    \end{tabular}
    }
    \small{Barnase and barstar (105 mutations, 14 neg., 91 pos.)}
    \vspace{0.35cm}
  \end{minipage}%
\end{table}

\vspace{-0.4cm}

\begin{table}[!bh]
  \centering
    \begin{minipage}{.49\textwidth}
    \resizebox{\textwidth}{!}{ \renewcommand{\arraystretch}{1.1}
    \small
    \begin{tabular}{l|ccc}
        \toprule
        Method & Spearman & Precision & Recall \\
        \midrule
        \textsc{Flex ddG} & 0.98 & 100\% & 100\% \\
        \midrule
        \textsc{GEMME} & \textbf{0.79} & \textbf{100}\% &	\textbf{80.00}\% \\
        \textsc{MSA Transformer} & 0.05 & \underline{66.67}\% &	40.00\% \\
        \textsc{ESM-IF} & 0.09 & 0.00\% & 0.00\% \\
        \textsc{RDE-Net.} & 0.15 & 50.00\% & 40.00\% \\
        \textsc{PPIformer} & \underline{0.34} & 60.00\% & \underline{60.00\%} \\
        \bottomrule
    \end{tabular}
    }
    \small{C.~thermophilum YTM1 and C.~thermophilum ERB1 (10 mutations, 5 neg., 5 pos.)}
    \end{minipage}%
  \hfill
  \begin{minipage}{.49\textwidth}
    \resizebox{\textwidth}{!}{ \renewcommand{\arraystretch}{1.1}
    \small
    \begin{tabular}{l|ccc}
        \toprule
        Method & Spearman & Precision & Recall \\
        \midrule
        \textsc{Flex ddG} & -0.05 & 29.27\% & 85.71\% \\
        \midrule
        \textsc{GEMME} & -0.10 & 0.00\% & 0.00\% \\
        \textsc{MSA Transformer} & \underline{0.09} & 0.00\% & 0.00\% \\
        \textsc{ESM-IF} & \textbf{0.10} & \underline{45.45\%} & \underline{71.43}\% \\
        \textsc{RDE-Net.} & -0.40 & 30.23\% & \textbf{92.86\%} \\
        \textsc{PPIformer} & 0.00 & \textbf{53.33\%} & 57.14\% \\
        \bottomrule
    \end{tabular}
    }
    \small{dHP1 Chromodomain and H3 tail (46 mutations, 14 neg., 2 zero, 30 pos.)}
  \end{minipage}%
\end{table}

\begin{table}[!bh]
    \centering
    \begin{minipage}{0.49\textwidth}
      \resizebox{\textwidth}{!}{ \renewcommand{\arraystretch}{1.1}
      \small
      \begin{tabular}{l|ccc}
          \toprule
          Method & Spearman & Precision & Recall \\
          \midrule
          \textsc{Flex ddG} & 0.29 & 44.44\% & 33.33\% \\
          \midrule
          \textsc{GEMME} & 0.20 & 0.00\% & 0.00\% \\
          \textsc{MSA-Transformer} & \underline{0.37} & 0.00\% & 0.00\% \\
          \textsc{ESM-IF} & 0.21 & 30.77\% & \textbf{33.33\%} \\
          \textsc{RDE-Net.} & 0.21 & \textbf{50.00\%} & \textbf{33.33\%} \\
          \textsc{PPIformer} & \textbf{0.43} & \underline{40.00\%} & \underline{16.67}\% \\
          \bottomrule
      \end{tabular}
      }
      \small{E6AP and UBCH7 (49 mutations, 12 neg., 37 pos.)\newline}
    \end{minipage}%
    \hfill
  \caption{Performance of \ppiformer and other competitive methods on 5 held-out test PPIs from the SKEMPI v2.0 dataset.}
  \label{fig:skempi-test}
\end{table}

%% file: assets/fig_idist_usalign.tex
\begin{figure}[b!]
  \centering
  \includegraphics[scale=0.2]{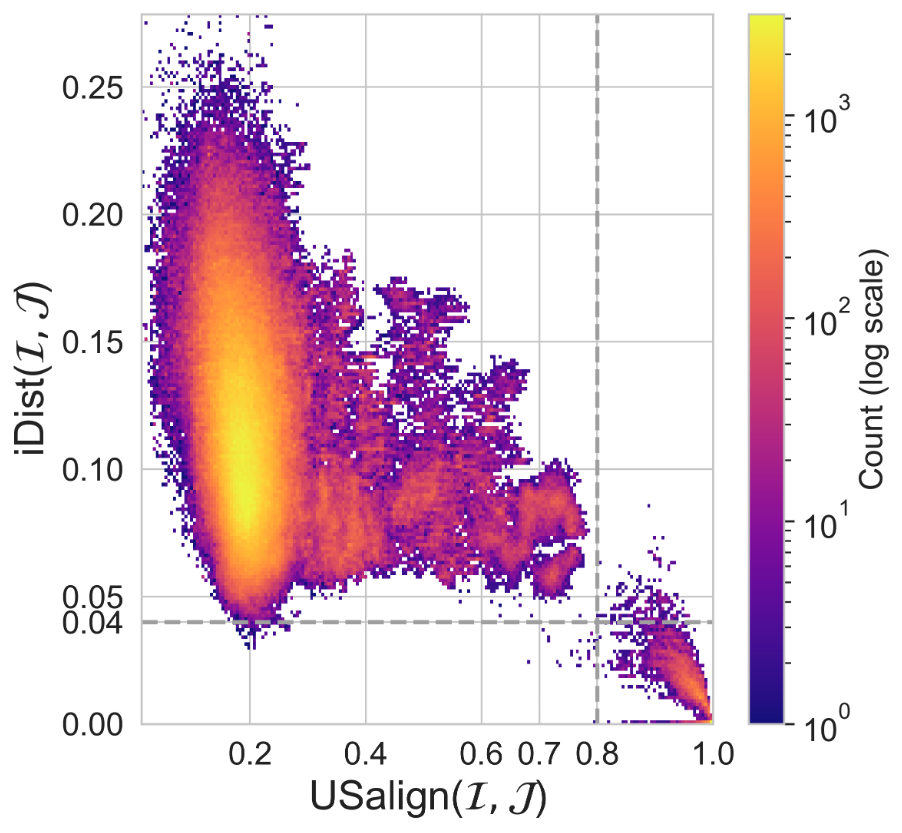}
  \caption[]{Extended panel for \Cref{fig:idist_vs_align} confirming that iDist accurately approximates not only iAlign but also independent \mbox{USalign} \citep{zhang2022us} approach for detecting near-duplicate PPI structures. See \Cref{appendix-data} for details.}
  \label{fig:idist_vs_usalign}
\end{figure}

%% file: iclr2024_conference.bbl
\begin{thebibliography}{76}
\providecommand{\natexlab}[1]{#1}
\providecommand{\url}[1]{\texttt{#1}}
\expandafter\ifx\csname urlstyle\endcsname\relax
  \providecommand{\doi}[1]{doi: #1}\else
  \providecommand{\doi}{doi: \begingroup \urlstyle{rm}\Url}\fi

\bibitem[Barlow et~al.(2018)Barlow, {\'O}~Conch{\'u}ir, Thompson, Suresh, Lucas, Heinonen, and Kortemme]{barlow2018flex}
Kyle~A. Barlow, Shane {\'O}~Conch{\'u}ir, Samuel Thompson, Pooja Suresh, James~E. Lucas, Markus Heinonen, and Tanja Kortemme.
\newblock {Flex ddG}: Rosetta ensemble-based estimation of changes in protein--protein binding affinity upon mutation.
\newblock \emph{The Journal of Physical Chemistry B}, 122\penalty0 (21):\penalty0 5389--5399, May 2018.
\newblock ISSN 1520-6106.
\newblock \doi{10.1021/acs.jpcb.7b11367}.
\newblock URL \url{https://doi.org/10.1021/acs.jpcb.7b11367}.

\bibitem[Berman et~al.(2000)Berman, Westbrook, Feng, Gilliland, Bhat, Weissig, Shindyalov, and Bourne]{berman2000protein}
Helen~M Berman, John Westbrook, Zukang Feng, Gary Gilliland, Talapady~N Bhat, Helge Weissig, Ilya~N Shindyalov, and Philip~E Bourne.
\newblock The protein data bank.
\newblock \emph{Nucleic acids research}, 28\penalty0 (1):\penalty0 235--242, 2000.
\newblock \doi{10.1093/nar/28.1.235}.
\newblock URL \url{https://www.ncbi.nlm.nih.gov/pmc/articles/PMC102472/}.

\bibitem[Burra et~al.(2009)Burra, Zhang, Godzik, and Stec]{burra2009global}
Prasad~V Burra, Ying Zhang, Adam Godzik, and Boguslaw Stec.
\newblock Global distribution of conformational states derived from redundant models in the pdb points to non-uniqueness of the protein structure.
\newblock \emph{Proceedings of the National Academy of Sciences}, 106\penalty0 (26):\penalty0 10505--10510, 2009.
\newblock \doi{10.1073/pnas.0812152106}.
\newblock URL \url{https://doi.org/10.1073/pnas.081215210}.

\bibitem[Bushuiev et~al.(2024)Bushuiev, Bushuiev, Sedlar, Pluskal, Damborsky, Mazurenko, and Sivic]{bushuiev2024revealing}
Anton Bushuiev, Roman Bushuiev, Jiri Sedlar, Tomas Pluskal, Jiri Damborsky, Stanislav Mazurenko, and Josef Sivic.
\newblock Revealing data leakage in protein interaction benchmarks.
\newblock In \emph{ICLR 2024 Workshop on Generative and Experimental Perspectives for Biomolecular Design}, 2024.
\newblock URL \url{https://openreview.net/forum?id=ORMXYUK5IY}.

\bibitem[Cheng et~al.(2015)Cheng, Zhang, and Brooks]{cheng2015pcalign}
Shanshan Cheng, Yang Zhang, and Charles~L Brooks.
\newblock {PCalign}: a method to quantify physicochemical similarity of protein-protein interfaces.
\newblock \emph{BMC bioinformatics}, 16\penalty0 (1):\penalty0 1--12, 2015.
\newblock \doi{10.1186/s12859-015-0471-x}.
\newblock URL \url{https://doi.org/10.1186/s12859-015-0471-x}.

\bibitem[Dauparas et~al.(2022)Dauparas, Anishchenko, Bennett, Bai, Ragotte, Milles, Wicky, Courbet, de~Haas, Bethel, et~al.]{dauparas2022robust}
Justas Dauparas, Ivan Anishchenko, Nathaniel Bennett, Hua Bai, Robert~J Ragotte, Lukas~F Milles, Basile~IM Wicky, Alexis Courbet, Rob~J de~Haas, Neville Bethel, et~al.
\newblock Robust deep learning--based protein sequence design using proteinmpnn.
\newblock \emph{Science}, 378\penalty0 (6615):\penalty0 49--56, 2022.
\newblock \doi{10.1126/science.add2187}.
\newblock URL \url{https://www.science.org/doi/10.1126/science.add2187}.

\bibitem[Dehouck et~al.(2013)Dehouck, Kwasigroch, Rooman, and Gilis]{dehouck2013beatmusic}
Yves Dehouck, Jean~Marc Kwasigroch, Marianne Rooman, and Dimitri Gilis.
\newblock {BeAtMuSiC}: prediction of changes in protein--protein binding affinity on mutations.
\newblock \emph{Nucleic acids research}, 41\penalty0 (W1):\penalty0 W333--W339, 2013.
\newblock \doi{10.1093/nar/gkt450}.
\newblock URL \url{https://pubmed.ncbi.nlm.nih.gov/23723246/}.

\bibitem[DeLano et~al.(2002)]{delano2002pymol}
Warren~L DeLano et~al.
\newblock {PyMOL}: An open-source molecular graphics tool.
\newblock \emph{CCP4 Newsl. Protein Crystallogr}, 40\penalty0 (1):\penalty0 82--92, 2002.
\newblock URL \url{https://pymol.org/}.

\bibitem[Devlin et~al.(2018)Devlin, Chang, Lee, and Toutanova]{devlin2018bert}
Jacob Devlin, Ming-Wei Chang, Kenton Lee, and Kristina Toutanova.
\newblock {BERT}: Pre-training of deep bidirectional transformers for language understanding.
\newblock \emph{arXiv preprint arXiv:1810.04805}, 2018.
\newblock \doi{10.48550/arXiv.1810.04805}.
\newblock URL \url{https://doi.org/10.48550/arXiv.1810.04805}.

\bibitem[Draizen et~al.(2022)Draizen, Murillo, Readey, Mura, and Bourne]{draizen2022prop3d}
Eli~J. Draizen, Luis Felipe~R. Murillo, John Readey, Cameron Mura, and Philip~E. Bourne.
\newblock {Prop3D}: A flexible, python-based platform for machine learning with protein structural properties and biophysical data.
\newblock \emph{bioRxiv}, 2022.
\newblock \doi{10.1101/2022.12.27.522071}.
\newblock URL \url{https://www.biorxiv.org/content/early/2022/12/30/2022.12.27.522071}.

\bibitem[Evans et~al.(2021)Evans, O’Neill, Pritzel, Antropova, Senior, Green, {\v{Z}}{\'\i}dek, Bates, Blackwell, Yim, et~al.]{evans2021protein}
Richard Evans, Michael O’Neill, Alexander Pritzel, Natasha Antropova, Andrew Senior, Tim Green, Augustin {\v{Z}}{\'\i}dek, Russ Bates, Sam Blackwell, Jason Yim, et~al.
\newblock Protein complex prediction with {AlphaFold-Multimer}.
\newblock \emph{biorxiv}, pp.\  2021--10, 2021.
\newblock \doi{10.1101/2021.10.04.463034}.
\newblock URL \url{https://doi.org/10.1101/2021.10.04.463034}.

\bibitem[Falcon \& {The PyTorch Lightning team}(2019)Falcon and {The PyTorch Lightning team}]{Falcon_PyTorch_Lightning_2019}
William Falcon and {The PyTorch Lightning team}.
\newblock {PyTorch Lightning}, Mar 2019.
\newblock URL \url{https://github.com/Lightning-AI/lightning}.

\bibitem[Feigin et~al.(2022)Feigin, Brainin, Norrving, Martins, Sacco, Hacke, Fisher, Pandian, and Lindsay]{feigin2022world}
Valery~L Feigin, Michael Brainin, Bo~Norrving, Sheila Martins, Ralph~L Sacco, Werner Hacke, Marc Fisher, Jeyaraj Pandian, and Patrice Lindsay.
\newblock {World Stroke Organization (WSO)}: global stroke fact sheet 2022.
\newblock \emph{International Journal of Stroke}, 17\penalty0 (1):\penalty0 18--29, 2022.
\newblock \doi{10.1177/17474930211065917}.
\newblock URL \url{https://pubmed.ncbi.nlm.nih.gov/34986727/}.

\bibitem[Fey \& Lenssen(2019)Fey and Lenssen]{fey2019fast}
Matthias Fey and Jan~Eric Lenssen.
\newblock Fast graph representation learning with {PyTorch Geometric}.
\newblock \emph{arXiv preprint arXiv:1903.02428}, 2019.
\newblock \doi{10.48550/arXiv.1903.02428}.
\newblock URL \url{https://doi.org/10.48550/arXiv.1903.02428}.

\bibitem[Gainza et~al.(2020)Gainza, Sverrisson, Monti, Rodola, Boscaini, Bronstein, and Correia]{gainza2020deciphering}
Pablo Gainza, Freyr Sverrisson, Frederico Monti, Emanuele Rodola, D~Boscaini, MM~Bronstein, and BE~Correia.
\newblock Deciphering interaction fingerprints from protein molecular surfaces using geometric deep learning.
\newblock \emph{Nature Methods}, 17\penalty0 (2):\penalty0 184--192, 2020.
\newblock \doi{10.1038/s41592-019-0666-6}.
\newblock URL \url{https://doi.org/10.1038/s41592-019-0666-6}.

\bibitem[Gainza et~al.(2023)Gainza, Wehrle, Van Hall-Beauvais, Marchand, Scheck, Harteveld, Buckley, Ni, Tan, Sverrisson, et~al.]{gainza2023novo}
Pablo Gainza, Sarah Wehrle, Alexandra Van Hall-Beauvais, Anthony Marchand, Andreas Scheck, Zander Harteveld, Stephen Buckley, Dongchun Ni, Shuguang Tan, Freyr Sverrisson, et~al.
\newblock De novo design of protein interactions with learned surface fingerprints.
\newblock \emph{Nature}, pp.\  1--9, 2023.
\newblock \doi{10.1038/s41586-023-05993-x}.
\newblock URL \url{https://doi.org/10.1038/s41586-023-05993-x}.

\bibitem[Ganea et~al.(2021)Ganea, Huang, Bunne, Bian, Barzilay, Jaakkola, and Krause]{ganea2021independent}
Octavian-Eugen Ganea, Xinyuan Huang, Charlotte Bunne, Yatao Bian, Regina Barzilay, Tommi Jaakkola, and Andreas Krause.
\newblock Independent {SE(3)}-equivariant models for end-to-end rigid protein docking.
\newblock \emph{arXiv preprint arXiv:2111.07786}, 2021.
\newblock \doi{10.48550/arXiv.2111.07786}.
\newblock URL \url{https://doi.org/10.48550/arXiv.2111.07786}.

\bibitem[Gao \& Skolnick(2010{\natexlab{a}})Gao and Skolnick]{gao2010ialign}
Mu~Gao and Jeffrey Skolnick.
\newblock {iAlign}: a method for the structural comparison of protein--protein interfaces.
\newblock \emph{Bioinformatics}, 26\penalty0 (18):\penalty0 2259--2265, 2010{\natexlab{a}}.
\newblock \doi{10.1093/bioinformatics/btq404}.
\newblock URL \url{https://doi.org/10.1093/bioinformatics/btq404}.

\bibitem[Gao \& Skolnick(2010{\natexlab{b}})Gao and Skolnick]{gao2010structural}
Mu~Gao and Jeffrey Skolnick.
\newblock Structural space of protein--protein interfaces is degenerate, close to complete, and highly connected.
\newblock \emph{Proceedings of the National Academy of Sciences}, 107\penalty0 (52):\penalty0 22517--22522, 2010{\natexlab{b}}.
\newblock \doi{10.1073/pnas.1012820107}.
\newblock URL \url{https://doi.org/10.1073/pnas.1012820107}.

\bibitem[Geng et~al.(2019{\natexlab{a}})Geng, Vangone, Folkers, Xue, and Bonvin]{geng2019isee}
Cunliang Geng, Anna Vangone, Gert~E Folkers, Li~C Xue, and Alexandre~MJJ Bonvin.
\newblock {iSEE}: Interface structure, evolution, and energy-based machine learning predictor of binding affinity changes upon mutations.
\newblock \emph{Proteins: Structure, Function, and Bioinformatics}, 87\penalty0 (2):\penalty0 110--119, 2019{\natexlab{a}}.
\newblock \doi{10.1002/prot.25630}.
\newblock URL \url{https://pubmed.ncbi.nlm.nih.gov/30417935/}.

\bibitem[Geng et~al.(2019{\natexlab{b}})Geng, Xue, Roel-Touris, and Bonvin]{geng2019finding}
Cunliang Geng, Li~C Xue, Jorge Roel-Touris, and Alexandre~MJJ Bonvin.
\newblock Finding the {$\Delta \Delta G$} spot: Are predictors of binding affinity changes upon mutations in protein--protein interactions ready for it?
\newblock \emph{Wiley Interdisciplinary Reviews: Computational Molecular Science}, 9\penalty0 (5):\penalty0 e1410, 2019{\natexlab{b}}.
\newblock \doi{10.1002/wcms.1410}.
\newblock URL \url{https://wires.onlinelibrary.wiley.com/doi/full/10.1002/wcms.1410}.

\bibitem[Hsu et~al.(2022)Hsu, Verkuil, Liu, Lin, Hie, Sercu, Lerer, and Rives]{hsu2022learning}
Chloe Hsu, Robert Verkuil, Jason Liu, Zeming Lin, Brian Hie, Tom Sercu, Adam Lerer, and Alexander Rives.
\newblock Learning inverse folding from millions of predicted structures.
\newblock \emph{bioRxiv}, 2022.
\newblock \doi{10.1101/2022.04.10.487779}.
\newblock URL \url{https://www.biorxiv.org/content/10.1101/2022.04.10.487779v2}.

\bibitem[Ivanov et~al.(2013)Ivanov, Khuri, and Fu]{ivanov2013targeting}
Andrei~A Ivanov, Fadlo~R Khuri, and Haian Fu.
\newblock Targeting protein--protein interactions as an anticancer strategy.
\newblock \emph{Trends in pharmacological sciences}, 34\penalty0 (7):\penalty0 393--400, 2013.
\newblock \doi{10.1016/j.tips.2013.04.007}.
\newblock URL \url{https://pubmed.ncbi.nlm.nih.gov/23725674/}.

\bibitem[Jamasb et~al.(2020)Jamasb, Vi{\~n}as, Ma, Harris, Huang, Hall, Li{\'o}, and Blundell]{jamasb2020graphein}
Arian~R Jamasb, Ramon Vi{\~n}as, Eric~J Ma, Charlie Harris, Kexin Huang, Dominic Hall, Pietro Li{\'o}, and Tom~L Blundell.
\newblock Graphein-a python library for geometric deep learning and network analysis on protein structures and interaction networks.
\newblock \emph{bioRxiv}, pp.\  2020--07, 2020.
\newblock \doi{10.1101/2020.07.15.204701}.
\newblock URL \url{https://doi.org/10.1101/2020.07.15.204701}.

\bibitem[Jankauskait{\.e} et~al.(2019)Jankauskait{\.e}, Jim{\'e}nez-Garc{\'\i}a, Dapk{\=u}nas, Fern{\'a}ndez-Recio, and Moal]{jankauskaite2019skempi}
Justina Jankauskait{\.e}, Brian Jim{\'e}nez-Garc{\'\i}a, Justas Dapk{\=u}nas, Juan Fern{\'a}ndez-Recio, and Iain~H Moal.
\newblock {SKEMPI 2.0}: an updated benchmark of changes in protein--protein binding energy, kinetics and thermodynamics upon mutation.
\newblock \emph{Bioinformatics}, 35\penalty0 (3):\penalty0 462--469, 2019.
\newblock \doi{10.1093/bioinformatics/bty635}.
\newblock URL \url{https://pubmed.ncbi.nlm.nih.gov/30020414/}.

\bibitem[Jin et~al.(2023)Jin, Sarkizova, Chen, Hacohen, and Uhler]{jin2023unsupervised}
Wengong Jin, Siranush Sarkizova, Xun Chen, Nir Hacohen, and Caroline Uhler.
\newblock Unsupervised protein-ligand binding energy prediction via neural euler's rotation equation.
\newblock \emph{arXiv preprint arXiv:2301.10814}, 2023.
\newblock \doi{10.48550/arXiv.2301.10814}.
\newblock URL \url{https://doi.org/10.48550/arXiv.2301.10814}.

\bibitem[Jing et~al.(2020)Jing, Eismann, Suriana, Townshend, and Dror]{jing2020learning}
Bowen Jing, Stephan Eismann, Patricia Suriana, Raphael~JL Townshend, and Ron Dror.
\newblock Learning from protein structure with geometric vector perceptrons.
\newblock \emph{arXiv preprint arXiv:2009.01411}, 2020.
\newblock \doi{10.48550/arXiv.2009.01411}.
\newblock URL \url{https://doi.org/10.48550/arXiv.2009.01411}.

\bibitem[Jumper et~al.(2021)Jumper, Evans, Pritzel, Green, Figurnov, Ronneberger, Tunyasuvunakool, Bates, {\v{Z}}{\'\i}dek, Potapenko, et~al.]{jumper2021highly}
John Jumper, Richard Evans, Alexander Pritzel, Tim Green, Michael Figurnov, Olaf Ronneberger, Kathryn Tunyasuvunakool, Russ Bates, Augustin {\v{Z}}{\'\i}dek, Anna Potapenko, et~al.
\newblock Highly accurate protein structure prediction with {AlphaFold}.
\newblock \emph{Nature}, 596\penalty0 (7873):\penalty0 583--589, 2021.
\newblock \doi{10.1038/s41586-021-03819-2}.
\newblock URL \url{https://doi.org/10.1038/s41586-021-03819-2}.

\bibitem[Karniadakis et~al.(2021)Karniadakis, Kevrekidis, Lu, Perdikaris, Wang, and Yang]{karniadakis2021physics}
George~Em Karniadakis, Ioannis~G Kevrekidis, Lu~Lu, Paris Perdikaris, Sifan Wang, and Liu Yang.
\newblock Physics-informed machine learning.
\newblock \emph{Nature Reviews Physics}, 3\penalty0 (6):\penalty0 422--440, 2021.
\newblock \doi{10.1038/s42254-021-00314-5}.
\newblock URL \url{https://www.nature.com/articles/s42254-021-00314-5}.

\bibitem[Kastritis \& Bonvin(2013)Kastritis and Bonvin]{kastritis2013binding}
Panagiotis~L Kastritis and Alexandre~MJJ Bonvin.
\newblock On the binding affinity of macromolecular interactions: daring to ask why proteins interact.
\newblock \emph{Journal of The Royal Society Interface}, 10\penalty0 (79):\penalty0 20120835, 2013.
\newblock \doi{10.1098/rsif.2012.0835}.
\newblock URL \url{https://doi.org/10.1098/rsif.2012.0835}.

\bibitem[Ketata et~al.(2023)Ketata, Laue, Mammadov, St{\"a}rk, Wu, Corso, Marquet, Barzilay, and Jaakkola]{ketata2023diffdock-pp}
Mohamed~Amine Ketata, Cedrik Laue, Ruslan Mammadov, Hannes St{\"a}rk, Menghua Wu, Gabriele Corso, C{\'e}line Marquet, Regina Barzilay, and Tommi~S Jaakkola.
\newblock {DiffDock-PP}: Rigid protein-protein docking with diffusion models.
\newblock \emph{arXiv preprint arXiv:2304.03889}, 2023.
\newblock \doi{10.48550/arXiv.2304.03889}.
\newblock URL \url{https://doi.org/10.48550/arXiv.2304.03889}.

\bibitem[Kingma \& Ba(2014)Kingma and Ba]{kingma2014adam}
Diederik~P Kingma and Jimmy Ba.
\newblock Adam: A method for stochastic optimization.
\newblock \emph{arXiv preprint arXiv:1412.6980}, 2014.
\newblock \doi{10.48550/arXiv.1412.6980}.
\newblock URL \url{https://doi.org/10.48550/arXiv.1412.6980}.

\bibitem[Laine et~al.(2019)Laine, Karami, and Carbone]{laine2019gemme}
Elodie Laine, Yasaman Karami, and Alessandra Carbone.
\newblock {GEMME}: a simple and fast global epistatic model predicting mutational effects.
\newblock \emph{Molecular biology and evolution}, 36\penalty0 (11):\penalty0 2604--2619, 2019.
\newblock \doi{10.1093/molbev/msz179}.
\newblock URL \url{https://doi.org/10.1093/molbev/msz179}.

\bibitem[Langan et~al.(2019)Langan, Boyken, Ng, Samson, Dods, Westbrook, Nguyen, Lajoie, Chen, Berger, et~al.]{langan2019novo}
Robert~A Langan, Scott~E Boyken, Andrew~H Ng, Jennifer~A Samson, Galen Dods, Alexandra~M Westbrook, Taylor~H Nguyen, Marc~J Lajoie, Zibo Chen, Stephanie Berger, et~al.
\newblock De novo design of bioactive protein switches.
\newblock \emph{Nature}, 572\penalty0 (7768):\penalty0 205--210, 2019.
\newblock \doi{10.1038/s41586-019-1432-8}.
\newblock URL \url{https://doi.org/10.1038/s41586-019-1432-8}.

\bibitem[Laroche et~al.(2000)Laroche, Heymans, Capaert, De~Cock, Demarsin, and Collen]{laroche2000recombinant}
Yves Laroche, Stephane Heymans, Sophie Capaert, Frans De~Cock, Eddy Demarsin, and D{\'e}sir{\'e} Collen.
\newblock Recombinant staphylokinase variants with reduced antigenicity due to elimination of b-lymphocyte epitopes.
\newblock \emph{Blood, The Journal of the American Society of Hematology}, 96\penalty0 (4):\penalty0 1425--1432, 2000.
\newblock \doi{10.1182/blood.V96.4.1425}.
\newblock URL \url{https://doi.org/10.1182/blood.V96.4.1425}.

\bibitem[Leman et~al.(2020)Leman, Weitzner, Lewis, Adolf-Bryfogle, Alam, Alford, Aprahamian, Baker, Barlow, Barth, et~al.]{leman2020rosetta}
Julia~Koehler Leman, Brian~D Weitzner, Steven~M Lewis, Jared Adolf-Bryfogle, Nawsad Alam, Rebecca~F Alford, Melanie Aprahamian, David Baker, Kyle~A Barlow, Patrick Barth, et~al.
\newblock Macromolecular modeling and design in rosetta: recent methods and frameworks.
\newblock \emph{Nature methods}, 17\penalty0 (7):\penalty0 665--680, 2020.
\newblock \doi{10.1038/s41592-020-0848-2}.
\newblock URL \url{https://doi.org/10.1038/s41592-020-0848-2}.

\bibitem[Liao \& Smidt(2022)Liao and Smidt]{liao2022equiformer}
Yi-Lun Liao and Tess Smidt.
\newblock Equiformer: Equivariant graph attention transformer for 3d atomistic graphs.
\newblock \emph{arXiv preprint arXiv:2206.11990}, 2022.
\newblock \doi{10.48550/arXiv.2206.11990}.
\newblock URL \url{https://doi.org/10.48550/arXiv.2206.11990}.

\bibitem[Liao et~al.(2023)Liao, Wood, Das, and Smidt]{liao2023equiformerv2}
Yi-Lun Liao, Brandon Wood, Abhishek Das, and Tess Smidt.
\newblock {EquiformerV2}: Improved equivariant transformer for scaling to higher-degree representations.
\newblock \emph{arXiv preprint arXiv:2306.12059}, 2023.
\newblock \doi{10.48550/arXiv.2306.12059}.
\newblock URL \url{https://doi.org/10.48550/arXiv.2306.12059}.

\bibitem[Lin et~al.(2023)Lin, Akin, Rao, Hie, Zhu, Lu, Smetanin, Verkuil, Kabeli, Shmueli, et~al.]{lin2023evolutionary}
Zeming Lin, Halil Akin, Roshan Rao, Brian Hie, Zhongkai Zhu, Wenting Lu, Nikita Smetanin, Robert Verkuil, Ori Kabeli, Yaniv Shmueli, et~al.
\newblock Evolutionary-scale prediction of atomic-level protein structure with a language model.
\newblock \emph{Science}, 379\penalty0 (6637):\penalty0 1123--1130, 2023.
\newblock \doi{10.1126/science.ade2574}.
\newblock URL \url{https://www.science.org/doi/10.1126/science.ade2574}.

\bibitem[Liu et~al.(2021)Liu, Luo, Li, Song, and Peng]{liu2021deep}
Xianggen Liu, Yunan Luo, Pengyong Li, Sen Song, and Jian Peng.
\newblock Deep geometric representations for modeling effects of mutations on protein-protein binding affinity.
\newblock \emph{PLoS computational biology}, 17\penalty0 (8):\penalty0 e1009284, 2021.
\newblock \doi{10.1371/journal.pcbi.1009284}.
\newblock URL \url{https://doi.org/10.1371/journal.pcbi.1009284}.

\bibitem[Lu et~al.(2020)Lu, Zhou, He, Jiang, Peng, Tong, and Shi]{lu2020recent}
Haiying Lu, Qiaodan Zhou, Jun He, Zhongliang Jiang, Cheng Peng, Rongsheng Tong, and Jianyou Shi.
\newblock Recent advances in the development of protein--protein interactions modulators: mechanisms and clinical trials.
\newblock \emph{Signal transduction and targeted therapy}, 5\penalty0 (1):\penalty0 213, 2020.
\newblock \doi{10.1038/s41392-020-00315-3}.
\newblock URL \url{https://doi.org/10.1038/s41392-020-00315-3}.

\bibitem[Luo et~al.(2023)Luo, Su, Wu, Su, Peng, and Ma]{luo2023rotamer}
Shitong Luo, Yufeng Su, Zuofan Wu, Chenpeng Su, Jian Peng, and Jianzhu Ma.
\newblock Rotamer density estimator is an unsupervised learner of the effect of mutations on protein-protein interaction.
\newblock In \emph{The Eleventh International Conference on Learning Representations}, 2023.
\newblock \doi{10.1101/2023.02.28.530137}.
\newblock URL \url{https://openreview.net/forum?id=_X9Yl1K2mD}.

\bibitem[Marchand et~al.(2022)Marchand, Van Hall-Beauvais, and Correia]{marchand2022computational}
Anthony Marchand, Alexandra~K Van Hall-Beauvais, and Bruno~E Correia.
\newblock Computational design of novel protein--protein interactions--an overview on methodological approaches and applications.
\newblock \emph{Current Opinion in Structural Biology}, 74:\penalty0 102370, 2022.
\newblock \doi{10.1016/j.sbi.2022.102370}.
\newblock URL \url{https://pubmed.ncbi.nlm.nih.gov/35405427/}.

\bibitem[Meier et~al.(2021)Meier, Rao, Verkuil, Liu, Sercu, and Rives]{meier2021language}
Joshua Meier, Roshan Rao, Robert Verkuil, Jason Liu, Tom Sercu, and Alex Rives.
\newblock Language models enable zero-shot prediction of the effects of mutations on protein function.
\newblock \emph{Advances in Neural Information Processing Systems}, 34:\penalty0 29287--29303, 2021.
\newblock \doi{10.1101/2021.07.09.450648}.
\newblock URL \url{https://doi.org/10.1101/2021.07.09.450648}.

\bibitem[Mirabello \& Wallner(2018)Mirabello and Wallner]{mirabello2018topology}
Claudio Mirabello and Bj{\"o}rn Wallner.
\newblock Topology independent structural matching discovers novel templates for protein interfaces.
\newblock \emph{Bioinformatics}, 34\penalty0 (17):\penalty0 i787--i794, 2018.
\newblock \doi{10.1093/bioinformatics/bty587}.
\newblock URL \url{https://academic.oup.com/bioinformatics/article/34/17/i787/5093254}.

\bibitem[Moal \& Fern{\'a}ndez-Recio(2012)Moal and Fern{\'a}ndez-Recio]{moal2012skempi}
Iain~H Moal and Juan Fern{\'a}ndez-Recio.
\newblock {SKEMPI}: a structural kinetic and energetic database of mutant protein interactions and its use in empirical models.
\newblock \emph{Bioinformatics}, 28\penalty0 (20):\penalty0 2600--2607, 2012.
\newblock \doi{10.1093/bioinformatics/bts489}.
\newblock URL \url{https://pubmed.ncbi.nlm.nih.gov/22859501/}.

\bibitem[Morehead et~al.(2021)Morehead, Chen, Sedova, and Cheng]{morehead2021dips}
Alex Morehead, Chen Chen, Ada Sedova, and Jianlin Cheng.
\newblock Dips-plus: The enhanced database of interacting protein structures for interface prediction.
\newblock \emph{arXiv preprint arXiv:2106.04362}, 2021.
\newblock \doi{10.48550/arXiv.2106.04362}.
\newblock URL \url{https://doi.org/10.48550/arXiv.2106.04362}.

\bibitem[Nikitin et~al.(2022)Nikitin, Mican, Toul, Bednar, Peskova, Kittova, Thalerova, Vitecek, Damborsky, Mikulik, et~al.]{nikitin2022computer}
Dmitri Nikitin, Jan Mican, Martin Toul, David Bednar, Michaela Peskova, Patricia Kittova, Sandra Thalerova, Jan Vitecek, Jiri Damborsky, Robert Mikulik, et~al.
\newblock Computer-aided engineering of staphylokinase toward enhanced affinity and selectivity for plasmin.
\newblock \emph{Computational and structural biotechnology journal}, 20:\penalty0 1366--1377, 2022.
\newblock \doi{10.1016/j.csbj.2022.03.004}.
\newblock URL \url{https://pubmed.ncbi.nlm.nih.gov/35386102/}.

\bibitem[Notin et~al.(2022)Notin, Dias, Frazer, Hurtado, Gomez, Marks, and Gal]{notin2022tranception}
Pascal Notin, Mafalda Dias, Jonathan Frazer, Javier~Marchena Hurtado, Aidan~N Gomez, Debora Marks, and Yarin Gal.
\newblock Tranception: protein fitness prediction with autoregressive transformers and inference-time retrieval.
\newblock In \emph{International Conference on Machine Learning}, pp.\  16990--17017. PMLR, 2022.
\newblock \doi{10.48550/arXiv.2205.13760}.
\newblock URL \url{https://doi.org/10.48550/arXiv.2205.13760}.

\bibitem[Pahari et~al.(2020)Pahari, Li, Murthy, Liang, Fragoza, Yu, and Alexov]{pahari2020saambe}
Swagata Pahari, Gen Li, Adithya~Krishna Murthy, Siqi Liang, Robert Fragoza, Haiyuan Yu, and Emil Alexov.
\newblock {SAAMBE-3D}: predicting effect of mutations on protein--protein interactions.
\newblock \emph{International journal of molecular sciences}, 21\penalty0 (7):\penalty0 2563, 2020.
\newblock \doi{10.3390/ijms21072563}.
\newblock URL \url{https://www.ncbi.nlm.nih.gov/pmc/articles/PMC7177817/}.

\bibitem[Paszke et~al.(2019)Paszke, Gross, Massa, Lerer, Bradbury, Chanan, Killeen, Lin, Gimelshein, Antiga, et~al.]{paszke2019pytorch}
Adam Paszke, Sam Gross, Francisco Massa, Adam Lerer, James Bradbury, Gregory Chanan, Trevor Killeen, Zeming Lin, Natalia Gimelshein, Luca Antiga, et~al.
\newblock {PyTorch}: An imperative style, high-performance deep learning library.
\newblock \emph{Advances in neural information processing systems}, 32, 2019.
\newblock \doi{10.48550/arXiv.1912.01703}.
\newblock URL \url{https://doi.org/10.48550/arXiv.1912.01703}.

\bibitem[Qasim et~al.(1997)Qasim, Ganz, Saunders, Bateman, James, and Laskowski]{qasim1997interscaffolding}
MA~Qasim, Philip~J Ganz, Charles~W Saunders, Katherine~S Bateman, Michael~NG James, and Michael Laskowski.
\newblock Interscaffolding additivity. association of p1 variants of eglin c and of turkey ovomucoid third domain with serine proteinases.
\newblock \emph{Biochemistry}, 36\penalty0 (7):\penalty0 1598--1607, 1997.
\newblock \doi{10.1021/bi9620870}.
\newblock URL \url{https://doi.org/10.1021/bi9620870}.

\bibitem[Rao et~al.(2021)Rao, Liu, Verkuil, Meier, Canny, Abbeel, Sercu, and Rives]{rao2021msa}
Roshan~M Rao, Jason Liu, Robert Verkuil, Joshua Meier, John Canny, Pieter Abbeel, Tom Sercu, and Alexander Rives.
\newblock Msa transformer.
\newblock In \emph{International Conference on Machine Learning}, pp.\  8844--8856. PMLR, 2021.
\newblock \doi{10.1101/2021.02.12.430858}.
\newblock URL \url{https://doi.org/10.1101/2021.02.12.430858}.

\bibitem[Remmert et~al.(2012)Remmert, Biegert, Hauser, and S{\"o}ding]{Remmert2012}
Michael Remmert, Andreas Biegert, Andreas Hauser, and Johannes S{\"o}ding.
\newblock {HHblits}: lightning-fast iterative protein sequence searching by {HMM-HMM} alignment.
\newblock \emph{Nature Methods}, 9\penalty0 (2):\penalty0 173--175, Feb 2012.
\newblock ISSN 1548-7105.
\newblock \doi{10.1038/nmeth.1818}.
\newblock URL \url{https://doi.org/10.1038/nmeth.1818}.

\bibitem[Ribeiro et~al.(2019)Ribeiro, R{\'\i}os-Vera, Melo, and Sch{\"u}ller]{ribeiro2019calculation}
Judemir Ribeiro, Carlos R{\'\i}os-Vera, Francisco Melo, and Andreas Sch{\"u}ller.
\newblock Calculation of accurate interatomic contact surface areas for the quantitative analysis of non-bonded molecular interactions.
\newblock \emph{Bioinformatics}, 35\penalty0 (18):\penalty0 3499--3501, 2019.
\newblock \doi{10.1093/bioinformatics/btz062}.
\newblock URL \url{https://doi.org/10.1093/bioinformatics/btz062}.

\bibitem[Rives et~al.(2021)Rives, Meier, Sercu, Goyal, Lin, Liu, Guo, Ott, Zitnick, Ma, et~al.]{rives2021biological}
Alexander Rives, Joshua Meier, Tom Sercu, Siddharth Goyal, Zeming Lin, Jason Liu, Demi Guo, Myle Ott, C~Lawrence Zitnick, Jerry Ma, et~al.
\newblock Biological structure and function emerge from scaling unsupervised learning to 250 million protein sequences.
\newblock \emph{Proceedings of the National Academy of Sciences}, 118\penalty0 (15):\penalty0 e2016239118, 2021.
\newblock \doi{10.1073/pnas.2016239118}.
\newblock URL \url{https://doi.org/10.1073/pnas.2016239118}.

\bibitem[Rodrigues et~al.(2021)Rodrigues, Pires, and Ascher]{rodrigues2021mmcsm}
Carlos~HM Rodrigues, Douglas~EV Pires, and David~B Ascher.
\newblock {mmCSM-PPI}: predicting the effects of multiple point mutations on protein--protein interactions.
\newblock \emph{Nucleic Acids Research}, 49\penalty0 (W1):\penalty0 W417--W424, 2021.
\newblock \doi{10.1093/nar/gkab273}.
\newblock URL \url{https://doi.org/10.1093/nar/gkab273}.

\bibitem[Scheller et~al.(2018)Scheller, Strittmatter, Fuchs, Bojar, and Fussenegger]{scheller2018generalized}
Leo Scheller, Tobias Strittmatter, David Fuchs, Daniel Bojar, and Martin Fussenegger.
\newblock Generalized extracellular molecule sensor platform for programming cellular behavior.
\newblock \emph{Nature chemical biology}, 14\penalty0 (7):\penalty0 723--729, 2018.
\newblock \doi{10.1038/s41589-018-0046-z}.
\newblock URL \url{https://doi.org/10.1038/s41589-018-0046-z}.

\bibitem[Schymkowitz et~al.(2005)Schymkowitz, Borg, Stricher, Nys, Rousseau, and Serrano]{schymkowitz2005foldx}
Joost Schymkowitz, Jesper Borg, Francois Stricher, Robby Nys, Frederic Rousseau, and Luis Serrano.
\newblock The {FoldX} web server: an online force field.
\newblock \emph{Nucleic acids research}, 33\penalty0 (suppl\_2):\penalty0 W382--W388, 2005.
\newblock \doi{10.1093/nar/gki387}.
\newblock URL \url{https://pubmed.ncbi.nlm.nih.gov/15980494/}.

\bibitem[Shan et~al.(2022)Shan, Luo, Yang, Hong, Su, Ding, Fu, Li, Chen, Ma, et~al.]{shan2022deep}
Sisi Shan, Shitong Luo, Ziqing Yang, Junxian Hong, Yufeng Su, Fan Ding, Lili Fu, Chenyu Li, Peng Chen, Jianzhu Ma, et~al.
\newblock Deep learning guided optimization of human antibody against {SARS-CoV-2} variants with broad neutralization.
\newblock \emph{Proceedings of the National Academy of Sciences}, 119\penalty0 (11):\penalty0 e2122954119, 2022.
\newblock \doi{10.1073/pnas.2122954119}.
\newblock URL \url{https://doi.org/10.1073/pnas.2122954119}.

\bibitem[Shin et~al.(2023{\natexlab{a}})Shin, Kumazawa, Imai, Hirokawa, and Kihara]{shin2023ppisurfer}
Woong-Hee Shin, Keiko Kumazawa, Kenichiro Imai, Takatsugu Hirokawa, and Daisuke Kihara.
\newblock Quantitative comparison of protein-protein interaction interface using physicochemical feature-based descriptors of surface patches.
\newblock \emph{Frontiers in Molecular Biosciences}, 10, 2023{\natexlab{a}}.
\newblock \doi{10.3389/fmolb.2023.1110567}.
\newblock URL \url{https://doi.org/10.3389/fmolb.2023.1110567}.

\bibitem[Shin et~al.(2023{\natexlab{b}})Shin, Kumazawa, Imai, Hirokawa, and Kihara]{shin2023quantitative}
Woong-Hee Shin, Keiko Kumazawa, Kenichiro Imai, Takatsugu Hirokawa, and Daisuke Kihara.
\newblock Quantitative comparison of protein-protein interaction interface using physicochemical feature-based descriptors of surface patches.
\newblock \emph{Frontiers in Molecular Biosciences}, 10:\penalty0 1110567, 2023{\natexlab{b}}.
\newblock \doi{10.3389/fmolb.2023.1110567}.
\newblock URL \url{https://doi.org/10.3389/fmolb.2023.1110567}.

\bibitem[Shroff et~al.(2019)Shroff, Cole, Morrow, Diaz, Donnell, Gollihar, Ellington, and Thyer]{shroff2019structure}
Raghav Shroff, Austin~W Cole, Barrett~R Morrow, Daniel~J Diaz, Isaac Donnell, Jimmy Gollihar, Andrew~D Ellington, and Ross Thyer.
\newblock A structure-based deep learning framework for protein engineering.
\newblock \emph{bioRxiv}, pp.\  833905, 2019.
\newblock \doi{10.1101/833905}.
\newblock URL \url{https://doi.org/10.1101/833905}.

\bibitem[Sora et~al.(2023)Sora, Laspiur, Degn, Arnaudi, Utichi, Beltrame, De~Menezes, Orlandi, Stoltze, Rigina, Sackett, Wadt, Schmiegelow, Tiberti, and Papaleo]{sora2023rosettaddg}
Valentina Sora, Adrian~Otamendi Laspiur, Kristine Degn, Matteo Arnaudi, Mattia Utichi, Ludovica Beltrame, Dayana De~Menezes, Matteo Orlandi, Ulrik~Kristoffer Stoltze, Olga Rigina, Peter~Wad Sackett, Karin Wadt, Kjeld Schmiegelow, Matteo Tiberti, and Elena Papaleo.
\newblock {RosettaDDGPrediction} for high-throughput mutational scans: From stability to binding.
\newblock \emph{Protein Science}, 32\penalty0 (1):\penalty0 e4527, 2023.
\newblock \doi{https://doi.org/10.1002/pro.4527}.
\newblock URL \url{https://onlinelibrary.wiley.com/doi/abs/10.1002/pro.4527}.

\bibitem[Steinegger \& S{\"o}ding(2017)Steinegger and S{\"o}ding]{steinegger2017mmseqs2}
Martin Steinegger and Johannes S{\"o}ding.
\newblock {MMseqs2} enables sensitive protein sequence searching for the analysis of massive data sets.
\newblock \emph{Nature biotechnology}, 35\penalty0 (11):\penalty0 1026--1028, 2017.
\newblock \doi{10.1038/nbt.3988}.
\newblock URL \url{https://doi.org/10.1038/nbt.3988}.

\bibitem[Szegedy et~al.(2016)Szegedy, Vanhoucke, Ioffe, Shlens, and Wojna]{szegedy2016rethinking}
Christian Szegedy, Vincent Vanhoucke, Sergey Ioffe, Jon Shlens, and Zbigniew Wojna.
\newblock Rethinking the inception architecture for computer vision.
\newblock In \emph{Proceedings of the IEEE conference on computer vision and pattern recognition}, pp.\  2818--2826, 2016.
\newblock \doi{10.48550/arXiv.1512.00567}.
\newblock URL \url{https://doi.org/10.48550/arXiv.1512.00567}.

\bibitem[Townshend et~al.(2019)Townshend, Bedi, Suriana, and Dror]{townshend2019end}
Raphael Townshend, Rishi Bedi, Patricia Suriana, and Ron Dror.
\newblock End-to-end learning on 3d protein structure for interface prediction.
\newblock \emph{Advances in Neural Information Processing Systems}, 32, 2019.
\newblock \doi{10.48550/arXiv.1807.01297}.
\newblock URL \url{https://doi.org/10.48550/arXiv.1807.01297}.

\bibitem[van Kempen et~al.(2023)van Kempen, Kim, Tumescheit, Mirdita, Lee, Gilchrist, S{\"o}ding, and Steinegger]{van2023fast}
Michel van Kempen, Stephanie~S Kim, Charlotte Tumescheit, Milot Mirdita, Jeongjae Lee, Cameron~LM Gilchrist, Johannes S{\"o}ding, and Martin Steinegger.
\newblock Fast and accurate protein structure search with foldseek.
\newblock \emph{Nature Biotechnology}, pp.\  1--4, 2023.
\newblock \doi{10.1038/s41587-023-01773-0}.
\newblock URL \url{https://doi.org/10.1038/s41587-023-01773-0}.

\bibitem[Vreven et~al.(2015)Vreven, Moal, Vangone, Pierce, Kastritis, Torchala, Chaleil, Jim{\'e}nez-Garc{\'\i}a, Bates, Fernandez-Recio, et~al.]{vreven2015updates}
Thom Vreven, Iain~H Moal, Anna Vangone, Brian~G Pierce, Panagiotis~L Kastritis, Mieczyslaw Torchala, Raphael Chaleil, Brian Jim{\'e}nez-Garc{\'\i}a, Paul~A Bates, Juan Fernandez-Recio, et~al.
\newblock Updates to the integrated protein--protein interaction benchmarks: docking benchmark version 5 and affinity benchmark version 2.
\newblock \emph{Journal of molecular biology}, 427\penalty0 (19):\penalty0 3031--3041, 2015.
\newblock \doi{10.1016/j.jmb.2015.07.016}.
\newblock URL \url{https://pubmed.ncbi.nlm.nih.gov/26231283/}.

\bibitem[Wang et~al.(2020)Wang, Cang, and Wei]{wang2020topology}
Menglun Wang, Zixuan Cang, and Guo-Wei Wei.
\newblock A topology-based network tree for the prediction of protein--protein binding affinity changes following mutation.
\newblock \emph{Nature Machine Intelligence}, 2\penalty0 (2):\penalty0 116--123, 2020.
\newblock \doi{10.1038/s42256-020-0149-6}.
\newblock URL \url{https://doi.org/10.1038/s42256-020-0149-6}.

\bibitem[Watson et~al.(2023)Watson, Juergens, Bennett, Trippe, Yim, Eisenach, Ahern, Borst, Ragotte, Milles, et~al.]{watson2023denovo}
Joseph~L Watson, David Juergens, Nathaniel~R Bennett, Brian~L Trippe, Jason Yim, Helen~E Eisenach, Woody Ahern, Andrew~J Borst, Robert~J Ragotte, Lukas~F Milles, et~al.
\newblock De novo design of protein structure and function with {RFdiffusion}.
\newblock \emph{Nature}, pp.\  1--3, 2023.
\newblock \doi{10.1038/s41586-023-06415-8}.
\newblock URL \url{https://www.nature.com/articles/s41586-023-06415-8}.

\bibitem[Xiong et~al.(2017)Xiong, Zhang, Zheng, and Zhang]{xiong2017bindprofx}
Peng Xiong, Chengxin Zhang, Wei Zheng, and Yang Zhang.
\newblock {BindProfX}: assessing mutation-induced binding affinity change by protein interface profiles with pseudo-counts.
\newblock \emph{Journal of molecular biology}, 429\penalty0 (3):\penalty0 426--434, 2017.
\newblock \doi{10.1016/j.jmb.2016.11.022}.
\newblock URL \url{https://pubmed.ncbi.nlm.nih.gov/27899282/}.

\bibitem[Yim et~al.(2023)Yim, Trippe, De~Bortoli, Mathieu, Doucet, Barzilay, and Jaakkola]{yim2023se}
Jason Yim, Brian~L Trippe, Valentin De~Bortoli, Emile Mathieu, Arnaud Doucet, Regina Barzilay, and Tommi Jaakkola.
\newblock {SE-(3)} diffusion model with application to protein backbone generation.
\newblock \emph{arXiv preprint arXiv:2302.02277}, 2023.
\newblock \doi{10.48550/arXiv.2302.02277}.
\newblock URL \url{https://doi.org/10.48550/arXiv.2302.02277}.

\bibitem[Zhang et~al.(2022)Zhang, Shine, Pyle, and Zhang]{zhang2022us}
Chengxin Zhang, Morgan Shine, Anna~Marie Pyle, and Yang Zhang.
\newblock {US-align}: universal structure alignments of proteins, nucleic acids, and macromolecular complexes.
\newblock \emph{Nature methods}, 19\penalty0 (9):\penalty0 1109--1115, 2022.
\newblock \doi{10.1038/s41592-022-01585-1}.
\newblock URL \url{https://doi.org/10.1038/s41592-022-01585-1}.

\bibitem[Zhang \& Skolnick(2004)Zhang and Skolnick]{zhang2004scoring}
Yang Zhang and Jeffrey Skolnick.
\newblock Scoring function for automated assessment of protein structure template quality.
\newblock \emph{Proteins: Structure, Function, and Bioinformatics}, 57\penalty0 (4):\penalty0 702--710, 2004.
\newblock \doi{10.1002/prot.20264}.
\newblock URL \url{https://pubmed.ncbi.nlm.nih.gov/15476259/}.

\bibitem[Zhang et~al.(2023)Zhang, Xu, Chenthamarakshan, Lozano, Das, and Tang]{zhang2023enhancing}
Zuobai Zhang, Minghao Xu, Vijil Chenthamarakshan, Aur{\'e}lie Lozano, Payel Das, and Jian Tang.
\newblock Enhancing protein language models with structure-based encoder and pre-training.
\newblock \emph{arXiv preprint arXiv:2303.06275}, 2023.
\newblock \doi{10.48550/arXiv.2303.06275}.
\newblock URL \url{https://doi.org/10.48550/arXiv.2303.06275}.

\end{thebibliography}
